\documentclass[10pt,journal,compsoc]{IEEEtran}
\usepackage{amsmath,amssymb,amsfonts}
\usepackage{algorithmic}
\usepackage{algorithm}
\usepackage{array}
\usepackage{booktabs}
\usepackage{caption}
\usepackage{multirow}
\usepackage{subcaption}
\usepackage{textcomp}
\usepackage{stfloats}
\usepackage{url}
\usepackage{verbatim}
\usepackage{graphicx}
\usepackage[pagebackref,breaklinks,colorlinks]{hyperref}
\usepackage[capitalize]{cleveref}
\usepackage{diagbox}
\usepackage{xcolor}
\usepackage{footnote}
\usepackage{bbding}

\ifCLASSOPTIONcompsoc
  \usepackage[nocompress]{cite}
\else
  \usepackage{cite}
\fi
\ifCLASSINFOpdf
\else
\fi
\begin{document}
%
\title{Global Learnable Attention for Single Image Super-Resolution}

\author{Jian-Nan~Su,
	   Min~Gan\IEEEauthorrefmark{1},~\IEEEmembership{Senior Member,~IEEE,}
        Guang-Yong~Chen,
	   Jia-Li~Yin,\\
        and~C. L. Philip Chen,~\IEEEmembership{Fellow,~IEEE}
\IEEEcompsocitemizethanks{\IEEEcompsocthanksitem * Corresponding author (E-mail: aganmin@aliyun.com)\protect\\
\IEEEcompsocthanksitem $\dagger$ https://github.com/laoyangui/DLSN\protect\\
\IEEEcompsocthanksitem Jian-Nan~Su, Min~Gan, Guang-Yong Chen and Jia-Li~Yin are with the College of Computer and Data Science, Fuzhou University, Fuzhou 350108, China.\protect\\
\IEEEcompsocthanksitem C. L. Philip Chen is with the School of Computer Science and Engineering, South China University of Technology, Guangzhou 510641, China, and also with the Faculty of Science and Technology, University of Macau, Macau, China
}}

\IEEEtitleabstractindextext{%
\begin{abstract}
Self-similarity is valuable to the exploration of non-local textures in single image super-resolution (SISR). Researchers usually assume that the importance of non-local textures is positively related to their similarity scores. In this paper, we surprisingly found that when repairing severely damaged query textures, some non-local textures with low-similarity which are closer to the target can provide more accurate and richer details than the high-similarity ones. In these cases, low-similarity does not mean inferior but is usually caused by different scales or orientations. Utilizing this finding, we proposed a Global Learnable Attention (GLA) to adaptively modify similarity scores of non-local textures during training instead of only using a fixed similarity scoring function such as the dot product. The proposed GLA can explore non-local textures with low-similarity but more accurate details to repair severely damaged textures. Furthermore, we propose to adopt Super-Bit Locality-Sensitive Hashing (SB-LSH) as a preprocessing method for our GLA. With the SB-LSH, the computational complexity of our GLA is reduced from quadratic to asymptotic linear with respect to the image size. In addition, the proposed GLA can be integrated into existing deep SISR models as an efficient general building block. Based on the GLA, we constructed a Deep Learnable Similarity Network (DLSN), which achieves state-of-the-art performance for SISR tasks of different degradation types (e.g. blur and noise). Our code and a pre-trained DLSN have been uploaded to GitHub\IEEEauthorrefmark{2} for validation.
\end{abstract}

}

\maketitle

\IEEEdisplaynontitleabstractindextext

%
\IEEEpeerreviewmaketitle

\IEEEraisesectionheading{\section{Introduction}\label{intro}}

%
%
%
%
\IEEEPARstart{S}{ingle} Image Super-Resolution (SISR) aims to recover a high-resolution (HR) image from its low-resolution (LR) input image, and it is useful in many important applications, such as medical imaging and object detection \cite{haris2021task,sui2022scan}. Formally, the degradation process of the HR image can be defined as 
\begin{equation}
\label{eq_hr_degradation}
\boldsymbol{I}=\boldsymbol{D}(\boldsymbol{H};\boldsymbol{\beta}),
\end{equation}
where $\boldsymbol{I}\in R^{h \times w}$ is an observed LR image and $\boldsymbol{H}\in R^{sh \times sw}$ is a corresponding HR image. $s$ is a down-scaling factor. $\boldsymbol{D}(\cdot)$ denotes the degradation operation with parameters $\boldsymbol{\beta}$ and the default choice for $\boldsymbol{D}(\cdot)$ in previous researches\cite{huang2015single,dong2016accelerating,lim2017enhanced,mei2021image} is the bicubic downsampling operator. If we consider the degradation  under non-ideal conditions (e.g. blurring and additive noise), which is closer to the real-world scenario, the degradation can be generalized as follows
\begin{equation}
\label{eq_hr_degradation_with_noise_blur}
\boldsymbol{I}=\boldsymbol{D}(\boldsymbol{k}*\boldsymbol{H};\boldsymbol{\beta}) + \boldsymbol{n},
\end{equation}
where $*$ denotes the convolution operation, $\boldsymbol{k}$ is a blur kernel with low pass filter, and $\boldsymbol{n}$ is an additive noise.

The SISR tasks, recovering $\boldsymbol{H}$ from $\boldsymbol{I}$, are regarded as severly ill-posed problems, particularly when the scaling factor is large. To generate visually pleasing results, it is common practice to use natural image priors such as the representative self-similarity. Thus, many self-similarity-based SISR methods \cite{ebrahimi2007solving,glasner2009super,huang2015single,mei2020image,mei2021image} were proposed to address the ill-posedness and achieved satisfactory reconstruction results. The self-similarity is based on an observation that small textures in a natural image tend to recur within and across scales of the image\cite{glasner2009super,zontak2011internal}. These repeated textures can provide valuable internal examples for a more faithful reconstruction. Essentially, the self-similarity provides valuable priors for exploring non-local image information. For example, when repairing hair textures, the related non-local hair regions are obviously more meaningful than low-frequency face regions or structured architectural regions. However, in SISR problems, previous studies usually assumed that the importance of non-local textures is positively related to their similarity scores, ignoring the flaws of this assumption, i.e., non-local textures with low-similarity may provide more accurate and richer details than the ones with 	high-similarity, especially when the query textures are severely damaged.

\begin{figure}[t]
  \centering
  \includegraphics[width=\linewidth]{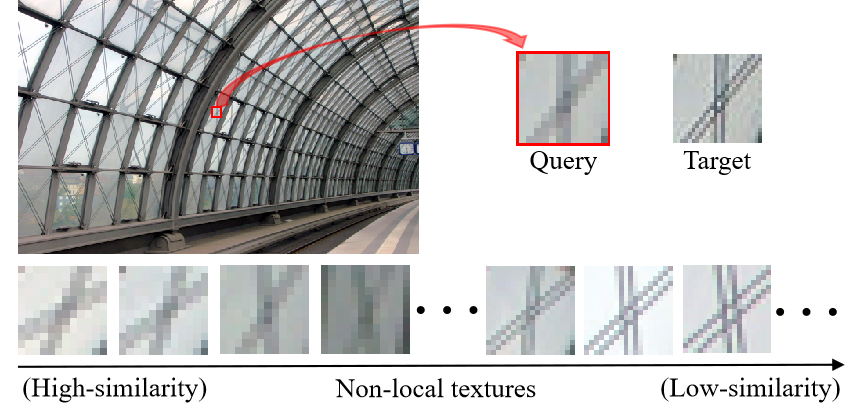}
   \caption{An illustration of the similarity-based non-local textures exploration. Non-local textures with low-similarity can provide more accurate details for SR reconstruction, especially when the query textures are severely damaged.}
   \vspace{-3mm}
   \label{fig_similarity_limitations}
\end{figure}

In this paper, we seek a deeper understanding of the role that the self-similarity plays in SISR tasks and expand its applicability in deep SISR models. In existing deep SISR models, the self-similarity is often integrated by non-local attention (NLA)\cite{wang2018non}, which was first proposed to explicitly model the long-range feature dependencies for high-level computer vision and also proven to be effective in SISR\cite{liu2018non,dai2019second,mei2020image}. These NLA-based SISR models achieved satisfactory reconstruction performance by utilizing the NLA to capture the self-similarity priors. However, the previous methods are all based on a basic assumption to explore the self-similarity: \emph{non-local textures that are more similar to the query textures can provide richer information}. We argue that the basic assumption is not always valid for SISR tasks. As illustrated in \cref{fig_similarity_limitations}, when a similarity-based non-local search is performed on the severely damaged query textures, non-local textures with high-similarity will get higher similarity scores. However, these high-similarity non-local textures obviously cannot provide the critical information for repairing the severely damaged query textures. Essentially, the reason for this defect is that the basic assumption of the self-similarity cannot handle the situation where low-similarity textures are more accurate and informative than high-similarity textures. Repairing such severely damaged textures is a critical and extremely difficult task for SISR, so it is meaningful to design a non-local textures exploration scheme that can handle this situation. Another problem of NLA-based SISR models is that the computational complexity is quadratic to the size of the input image, which is generally unacceptable for SISR tasks.

\begin{figure*}[!htbp]
  \centering
  \includegraphics[width=0.85\linewidth,height=0.45\linewidth]{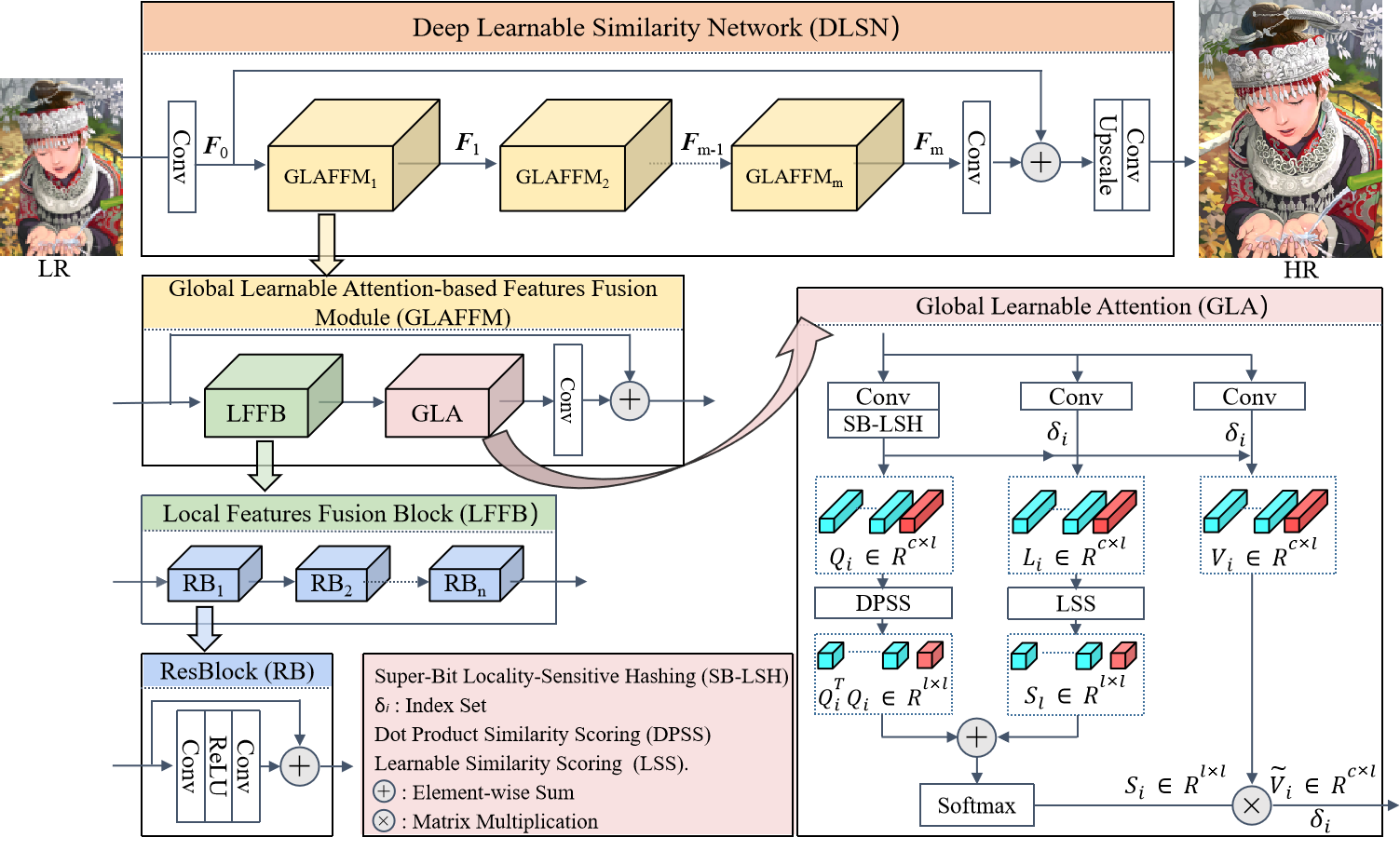}
   \caption{An illustration of our DLSN. For the convenience of description, we only show the process of \emph{i}-th bucket in GLA and the implementation details of the process can be found in \cref{alg:gla}.} 
  \vspace{-3mm}
   \label{fig_dlsn}
\end{figure*}

To address these issues, we proposed a Global Learnable Attention (GLA) module, which can modify similarity score between two features adaptively during training and has asymptotic linear computational complexity with respect to the size of the input image. Our GLA module consists of two core parts: a Learnable Similarity Scoring (LSS) function and a Super-Bit Locality-Sensitive Hashing (SB-LSH)\cite{ji2012super}. With the SB-LSH, our GLA module can perform hashing on input features and compute similarity only within the same hash bucket. The SB-LSH not only can reduce the computational complexity of the standard self-attention from quadratic to asymptotic linear, but also eliminate a large amount of redundant and irrelevant non-local information. The proposed LSS function aims to address the inherent limitations of the self-similarity (see \cref{fig_similarity_limitations}) by providing trainable parameters to adaptively modify similarity scores between input features. Furthermore, we constructed a Global Learnable Attention based Features Fusion Module (GLAFFM) to fuse local adjacency information and learnable non-local self-similarity information. The local adjacency information is captured by the common component of Local Features Fusion Block (LFFB) that consists of several simple residual convolution layers. Finally, we integrated some GLAFFMs into a residual backbone to build our Deep Learnable Similarity Network (DLSN) (see \cref{fig_dlsn}) for SISR tasks.

The effectiveness of our DLSN will be verified in the experiment section for different degradation types of SISR tasks: bicubic-downscale, blur-downscale, and noisy-downscale. In all degradation types, our DLSN outperforms other state-of-the-art SISR models \cite{zhang2018residual,zhang2018image,dai2019second,mei2020image} by a large margin. In addition, to verify the universality of our GLA in deep SISR, we integrated our GLA into some representative deep SISR models, such as FSRCNN \cite{dong2016accelerating}, EDSR\cite{lim2017enhanced}, and RCAN\cite{zhang2018image}. Experimental results demonstrate that our GLA can significantly improve the SR performance of various deep SISR models. Ablation studies are also conducted to analyze the impact of the proposed LFFB and GLA on the reconstruction results.

The main contributions of this paper are summarized as follows.
\begin{itemize}
\item{We provide some new insights into the self-similarity-based SISR solutions and argue that the basic assumption of the self-similarity is not flexible enough for SISR tasks. (As shown in \cref{fig_similarity_limitations})}
\item {The proposed Global Learnable Attention (GLA) with Super-Bit Locality-Sensitive Hashing (SB-LSH) can adaptively modify the similarity scores during training and has asymptotic linear computational complexity.}
\item {A new Deep Learnable Similarity Network (DLSN) is designed by using our Global Learnable Attention-based Features Fusion Modules (GLAFFMs) and achieves the state-of-the-art SR performance both quantitatively and qualitatively.}
\end{itemize}

\section{Related Work} \label{related_work}
The self-similarity has been widely applied to image generation problems. It assumes that similar texture patterns frequently recur within and across scales of the same image. Many classical SISR methods achieve satisfactory SR performance by exploring the self-similarity\cite{ebrahimi2007solving,protter2008generalizing,glasner2009super,freedman2011image,huang2015single}. The difference between these classical SISR methods in exploring self-similarity is mainly in the range of non-local search space. For example, to reduce the computational complexity of exploring the self-similarity, the search space is usually constrained to some specified local regions\cite{protter2008generalizing,freedman2011image}. To achieve more accurate reconstruction quality, researchers usually extend the search space of the self-similarity to cross-scale images\cite{ebrahimi2007solving,glasner2009super}. In addition, the search space can be further expanded by modeling geometric transformations\cite{huang2015single}. 

Although these classical SISR methods differ in the range of the self-similarity search space, they all follow the same principle of using the self-similarity, which assigns larger weights to non-local textures that are more similar to the query textures. This principle is still used in deep SISR models for exploring the self-similarity information\cite{liu2018non,dai2019second,mei2020image,zhou2020cross,mei2021image}. Specifically, these deep SISR models use cosine similarity to measure the similarity scores between two features and assign higher weights to features with high-similarity. However, as illustrated in \cref{fig_similarity_limitations}, we argue that this principle is not reasonable enough for SISR tasks. This motivates us to explore a learnable similarity scoring method to address this drawback.

In deep learning-based SISR, the non-local attention (NLA)\cite{wang2018non} is commonly used to explore the self-similarity. However, the computational complexity of the NLA is quadratic to the size of the input image \cite{liu2018non,dai2019second,mei2020image}, which significantly affects the application of the NLA in deep SISR. Fortunately, there are many approaches to reduce the computational complexity of the NLA by using sparse attention, such as locality sensitive hashing attention\cite{kitaev2019reformer,mei2021image}, routing attention\cite{roy2021efficient}, and BigBird\cite{zaheer2020big}. For example, in routing attention\cite{roy2021efficient}, the query feature is routed to a limited number of context elements through its cluster assigned by the spherical k-means clustering. Compared with the standard NLA, these sparse attention methods can reduce the complexity significantly by making each feature interact with less but more similar features. Motivated by sparse attention, we propose to adopt the Super-Bit Locality-Sensitive Hashing (SB-LSH)\cite{ji2012super} as a preprocessing method of our Global Learnable Attention (GLA). The reason we used SB-LSH to provide sparse attention is that it has two advantages: (1) It only adds negligible extra computation. (2) It is theoretically guaranteed \cite{ji2012super} that the SB-LSH can achieve a small hashing variance.

\section{Methodology}
In this section, we will introduce our Deep Learnable Similarity Network (DLSN) in details. The DLSN consists of a residual backbone with our Global Learnable Attention-based Features Fusion Modules (GLAFFMs). For SISR tasks, there are two kinds of information which are critical for improving the reconstruction results: local adjacency information and learnable non-local self-similarity information. The GLAFFM is designed for fusing these two types of critical information 	captured by the Local Features Fusion Block (LFFB) and our Global Learnable Attention (GLA), respectively. As discussed in most previous studies, we used the LFFB as a common component to capture the local adjacency information. We start with an overview of the proposed DLSN and then introduce the details of each component in the GLAFFM.

\subsection{Overview of DLSN} \label{dlsn}
As shown in \cref{fig_dlsn}, our DLSN is an end-to-end SR framework which is composed of three parts: low-resolution (LR) features extraction, local and global deep features fusion, and HR image reconstruction. As suggested in \cite{lim2017enhanced,zhang2018image}, only one convolutional layer with trainable parameters $\boldsymbol{\alpha}$ is used to extract the shallow feature $\boldsymbol{F}_0$ from the given LR image $\boldsymbol{I}$. This procedure can be formally defined as follows
\begin{equation}
\boldsymbol{F}_0=\Phi(\boldsymbol{I};\boldsymbol{\alpha}),
  \label{eq_shallow_feature_extract}
\end{equation}
where $\Phi(\cdot)$ is the convolution operation in LR features extraction part. Then, $\boldsymbol{F}_0$ is fed into the local and global deep features fusion part with $m$ GLAFFMs to obtain refined deep features $\boldsymbol{F}_m$
\begin{equation}
\boldsymbol{F}_m=\Psi(\boldsymbol{F}_0;\boldsymbol{\beta}),
  \label{eq_deep_feature_extract}
\end{equation}
where $\Psi(\cdot)$ represents the function of our local and global deep feature fusion part with trainable parameters $\boldsymbol{\beta}$. Finally, we upscale the obtained deep features $\boldsymbol{F}_m$ by sub-pixel convolution $\uparrow$ \cite{shi2016real} and then use it to generate a high-resolution image $\boldsymbol{\hat{I}}$ in the HR image reconstruction part as
\begin{equation}
\boldsymbol{\hat{I}}=\Omega(\boldsymbol{F}_m\uparrow;\boldsymbol{\gamma}),
  \label{eq_hr_image_reconstruction}
\end{equation}
where $\Omega(\cdot)$ denotes the HR image reconstruction part, which is implemented by a convolutional layer of 3 filters with trainable parameters $\boldsymbol{\gamma}$ for the final RGB image reconstruction. We can combine the three parts above in our DLSN as
\begin{equation}
\boldsymbol{\hat{I}}=\text{DLSN}(\boldsymbol{I};(\boldsymbol{\alpha},\boldsymbol{\beta},\boldsymbol{\gamma},\uparrow)),
  \label{eq_dlsn}
\end{equation}
where $\text{DLSN}(\cdot)$ is the function of our deep learnable similarity network. In addition, to focus on  learning high-frequency information and preventing gradients from exploding, we use a long skip connection in the DLSN to directly bypass abundant low-frequency information.

\subsection{Global Learnable Attention-based Features Fusion Module (GLAFFM)} \label{glaffm}
The structure of GLAFFM is shown in \cref{fig_dlsn}, from which we can see that the GLAFFM serves as a basic module of our DLSN. Specifically, each GLAFFM is also with the residual architecture and consists of a Local Features Fusion Block (LFFB), a Global Learnable Attention (GLA) and a feature refinement convlution layer. The LFFB is responsible for capturing locality inductive bias, while the GLA is exploring global information. 

The trainable parameters of GLAFFM are omitted for simplicity and the corresponding function in the \emph{i}-th GLAFFM can be defined as
\begin{equation}
\boldsymbol{F}_i=\Psi_i(\boldsymbol{F}_{i-1})=\Psi_i(\Psi_{i-1}(\cdots \Psi_2(\Psi_1(\boldsymbol{F}_0)))),
\label{eq_glaffm}
\end{equation}
where $\Psi_i$ represents the function of \emph{i}-th GLAFFM. $\boldsymbol{F}_{i-1}$ and $\boldsymbol{F}_{i}$ are the input and the output of the \emph{i}-th GLAFFM. 

\subsubsection{Local features fusion block (LFFB)} \label{lffb}
As discussed in most previous studies \cite{lim2017enhanced,anwar2022densely}, we used the LFFB to exploit the locality of the convolution for capturing the local information of nature images. The LFFB is the cornerstone of SR reconstruction, which is formed by stacking $n$ simplified residual blocks (see \cref{fig_dlsn} for more details).

\subsubsection{Global learnable attention (GLA)} \label{nllsa}
Our GLA can explore global information by summarizing related features from input feature maps. Given the input feature maps $\boldsymbol{X}\in R^{h\times w\times c}$, we reshap it into a 1-D feature $\boldsymbol{X^{'}}\in R^{hw\times c}$ for illustration purposes. Then, the attention process of the query feature vector $\boldsymbol{x}_i$ can be generally formulated as
\begin{equation}
  f(\boldsymbol{x}_i)=\sum_{j=1}^{n}\frac{{\rm exp}(s(\boldsymbol{x}_i,\boldsymbol{x}_j))}{\sum_{k=1}^{n}{\rm exp}(s(\boldsymbol{x}_i,\boldsymbol{x}_k))}\phi_v(\boldsymbol{x}_j),
  \label{eq_pixel_wise_similarity}
\end{equation}
where $n=hw$, $\boldsymbol{x}_j$ and $\boldsymbol{x}_k$ are the \emph{j}-th and \emph{k}-th feature vectors on $\boldsymbol{X^{'}}$ respectively. $\phi_v(\cdot)$ is a feature transformation function that generates value vectors. $s(\cdot , \cdot )$ is used to measure similarity and composes a learnable similarity scoring function $s_l(\boldsymbol{x}_i)$ and a fixed dot product similarity scoring function $s_f(\boldsymbol{x}_i,\boldsymbol{x}_j)$ 
\begin{equation}
s(\boldsymbol{x}_i,\boldsymbol{x}_j)=s^{j}_l(\boldsymbol{x}_i)+s_f(\boldsymbol{x}_i,\boldsymbol{x}_j),
  \label{eq_learnable_fc_and_fixed_dot_product}
\end{equation}
where $s^{j}_l(\cdot)$ represents the \emph{j}-th component in $s_l(\cdot)$, and $s_f(\cdot,\cdot)$ is the standard fixed dot product similarity scoring function, which can be defined as
\begin{equation}
s_f(\boldsymbol{x}_i,\boldsymbol{x}_j)=\phi_q(\boldsymbol{x}_i)^\mathrm{T}\phi_k(\boldsymbol{x}_j),
  \label{eq_fixed_dot_product}
\end{equation}
where $\phi_q(\cdot)$ and $\phi_k(\cdot)$ are feature transformations which we use to generate query and key vectors. In practice, our GLA shared the parameters in $\phi_q(\cdot)$ and $\phi_k(\cdot)$ to reduce the computational cost, and we found that this modification didn't reduce the SR performance.

Existing non-local deep SISR methods\cite{liu2018non,dai2019second,mei2020image,mei2021image} assume $s(\cdot,\cdot)=s_f(\cdot,\cdot)$, and we have discussed the limitations in Section \ref{intro}. To overcome the limitations, we proposed a learnable similarity scoring function $s_l(\cdot)$ to adaptively modify similarity scores. It is worth noting that the similarity scoring function is highly nonlinear and varies from object to object. Here we used a single hidden layer feedforward neural network (FNN) that has achieved excellent representation ability to revise the fixed dot product similarity scoring. In addition, the modified term learned from the transformed feature $s_l(\cdot)$ directly through the FNN with a single hidden layer (deeper networks are also possible) has linear computational complexity with respect to the size of the input image. Formally, the learnable similarity scoring function $s_l(\cdot)$ can be defined as
\begin{equation}
s_l(\boldsymbol{x}_i)=\boldsymbol{W}_2\sigma(\boldsymbol{W}_1\phi_l(\boldsymbol{x}_i)+\boldsymbol{b}_1)+\boldsymbol{b}_2,
  \label{eq_learnable_fc}
\end{equation}
where $\sigma(\cdot)$ is the ReLU activation and $\boldsymbol{x}_i\in R^{c}$, $\boldsymbol{W}_1\in R^{n\times c}$, $\boldsymbol{b}_1\in R^{n}$,$\boldsymbol{W}_2\in R^{n\times n}$, $\boldsymbol{b}_2\in R^{n}$. 

\subsection{Super-Bit Locality-Sensitive Hashing (SB-LSH)} \label{sb_lsh}
As mentioned in Section \ref{related_work}, the sparse attention has been widely used in deep learning\cite{kitaev2019reformer,mei2021image,roy2021efficient,zaheer2020big} to reduce the computational complexity of the standard non-local attention from quadratic to asymptotic linear by computing the similarity only within each bucket. In this paper, we propose to adopt the SB-LSH\cite{ji2012super} to hash global related features into the bucket of the query feature. The SB-LSH provides an estimation of angular similarity with negligible extra computation and shows that the more similar elements are more likely to fall into the same hash bucket. Thus, the SB-LSH can be used as a suitable preprocessing method for our Global Learnable Attention (GLA). Given $b$ hash buckets, we first project the query feature $\boldsymbol{x}_i$ with dimension $c$ onto an orthonormal basis $\boldsymbol{M}\in R^{b\times c}$:
\begin{equation}
\boldsymbol{x}_i^{'}=\boldsymbol{M}\boldsymbol{x}_i.
  \label{eq_ortho_project}
\end{equation}

Then, the assigned hash bucket of $\boldsymbol{x}_i$ can be expressed as $h(\boldsymbol{x}_i^{'})= \rm argmax(\boldsymbol{x}_i^{'})$, where $\rm argmax(\cdot)$ finds the index that gives the maximum value from $\boldsymbol{x}_i^{'}$. Finally, global features which are assigned in the same hash bucket $\lambda_i$ with the query feature $\boldsymbol{x_i}$ can be defined as
\begin{equation}
\lambda_i=\{\boldsymbol{x}_j|h(\boldsymbol{x}_i^{'})=h(\boldsymbol{x}_j^{'})\}.
  \label{eq_query_hash_buckets}
\end{equation}

The SB-LSH is simultaneously performed for all query features with batch matrix multiplication, which only adds negligible extra computation. With the preprocessing method SB-LSH, our GLA can achieve asymptotic linear computational complexity with respect to the size of the input image by computing the non-local attention only within the same hash bucket. Formally, the GLA with the preprocessing method SB-LSH can be derived from Eq. \eqref{eq_pixel_wise_similarity} and Eq. \eqref{eq_query_hash_buckets} as follows
\begin{equation}
  f(\boldsymbol{x}_i)=\sum_{\boldsymbol{x}_j \in \lambda_i}\frac{{\rm exp}(s(\boldsymbol{x}_i,\boldsymbol{x}_j))}{\sum_{\boldsymbol{x}_k \in \lambda_i}{\rm exp}(s(\boldsymbol{x}_i,\boldsymbol{x}_k))}\phi_v(\boldsymbol{x}_j),
  \label{eq_hash_pixel_wise_similarity}
\end{equation}
where $\lambda_i$ contains the features which are assigned in the same hash bucket with the query feature $\boldsymbol{x}_i$. Next, we will introduce some solutions to the problems which arise when using the SB-LSH. 

\noindent{\bf{Uneven bucket size.}} In practice, the size of hash buckets tend to be uneven, which makes it difficult to perform parallel computing. To solve the problems, we sort all query features by their bucket number and the sorted permutation is denoted as $\xi:i\rightarrow \xi(i)$, where $\xi(i)$ is the new position corresponding to the original index $i$. Given buckets size $l$, the sorted features in the \emph{k}-th chunk $C_k$ can be expressed as
\begin{equation}
C_k=\{\boldsymbol{x}_{lk+1},\boldsymbol{x}_{lk+2},...,\boldsymbol{x}_{l(k+1)}\}.
  \label{eq_k_th_chunk_features}
\end{equation}

Finally, the hash bucket $\lambda_i$ of the query feature $\boldsymbol{x}_i$ can be updated as
\begin{equation}
\lambda_i=C_k,
  \label{eq_update_att_bucket}
\end{equation}
where the sorted index $\xi(i)$ of $\boldsymbol{x}_i$ is between $lk+1$ and $l(k+1)$. We can now obtain hash buckets of the same size for parallel computing. Some new hash buckets may cross the original bucket boundaries, which can be solved by extending the attention over adjacent buckets. After using adjacent buckets, the search space of non-local features is expanded to $3l$.

\noindent{\bf{Multi-hash SB-LSH.}} The hash operation based on SB-LSH has a small probability that similar features are not split into the same bucket. This probability can be reduced by merging the results from multiple rounds of hashing. In the \emph{r}-th round of hashing, the result of the query feature $\boldsymbol{x}_i$ processed by our GLA (see Eq. \eqref{eq_hash_pixel_wise_similarity}) is defined as $f_{r}(\boldsymbol{x}_i)$. Then, the results of our GLA with multi-hash SB-LSH is regarded as the weighted sum of each hashing attention result. Formally, the multi-hash GLA results $\boldsymbol{\hat{x}}_i$ can be obtained by
\begin{equation}
\boldsymbol{\hat{x}}_i=\sum_{r}\omega_rf_{r}(\boldsymbol{x}_i),
  \label{eq_multi_hash_gla}
\end{equation}
where $\omega_r$ is the weight coefficient of \emph{r}-th round hashing. The $\omega_r$ represents the normalized similarity between the query feature and all features in its assigned \emph{r}-th round bucket
\begin{equation}
\omega_r=\frac{\sum_{\boldsymbol{x}_j \in \lambda^{r}_i} s(\boldsymbol{x}_i,\boldsymbol{x}_j)}{\sum_{\hat{r}=1}^{h}\sum_{\boldsymbol{x}_k \in \lambda^{\hat{r}}_i} s(\boldsymbol{x}_i,\boldsymbol{x}_k)},
  \label{eq_r_round_weight}
\end{equation}
where $\lambda^{r}_i$ is a set of global features which are assigned in the same hash bucket with the query feature $\boldsymbol{x}_i$ in the \emph{r}-th round of hashing. $h$ is the number of hashing rounds. The effectiveness of the multi-hash SB-LSH will be verified in the experiment section.

More implementation details of our GLA with SB-LSH can be found in \cref{alg:gla}. For illustration, we only show the process of the \emph{i}-th bucket in GLA, the remaining buckets are executed in parallel in the same way.

\subsection{Computational Complexity of Similarity Scoring}
Given an input feature $\boldsymbol{X}\in R^{h\times w\times c}$, the standard non-local attention on $\boldsymbol{X}$ is $O((hw)^2c)$ in computational complexity. After hashing the input feature $\boldsymbol{X}$ into $\frac{hw}{l}$ buckets with size $l$, the computational complexity of the standard fixed dot product similarity scoring $s_f(\cdot,\cdot)$ and our learnable similarity scoring function $s_l(\cdot)$ are $O(hwlc)$ and $O(hwlc+hwl^2)$, respectively. Thus, the computational cost of our GLA in calculating the similarity scores is $O(2hwlc+hwl^2)$. It means that our GLA with SB-LSH has asymptotic linear computational complexity with respect to the size of the input image.

 \begin{algorithm}[!htbp]
\caption{Global Learnable Attention (GLA) with Super-Bit Locality-Sensitive Hashing (SB-LSH).}\label{alg:gla}
\begin{algorithmic}[1]
\STATE \textbf{Input:} input features $\boldsymbol{X}\in R^{c\times h\times w}$.
\STATE $\boldsymbol{Q} \gets {\rm {Conv}}(\boldsymbol{X})$\ \ \textcolor{gray}{\# convolution with kernel size 3.}
\STATE $\boldsymbol{Q} \gets Reshape(\boldsymbol{Q})$ \ \ \textcolor{gray}{\# reshape $\boldsymbol{Q}$ to $R^{c\times hw}$.}
\STATE {\textsc{SB-LSH}}$(\boldsymbol{q}_i)$ \ \ \textcolor{gray}{\# $\boldsymbol{q}_i\in R^c$ is the \emph{i}-th component of $\boldsymbol{Q}$.}
\STATE \hspace{0.32cm}randomly initialize a matrix $\boldsymbol{H}\in R^{b\times c}$
\STATE \hspace{0.32cm}$\boldsymbol{M} \gets$ orthogonalize $\boldsymbol{H}$ via Gram-Schmidt process
\STATE \hspace{0.32cm}$\boldsymbol{q}_i^{'} \gets \boldsymbol{M}\boldsymbol{q}_i$
\STATE \hspace{0.32cm}${\rm index} \gets {\rm{argmax}}(\boldsymbol{q}_i^{'})$
\STATE \hspace{0.32cm}$\boldsymbol{Q}_i \gets\{\}$\ \textcolor{gray}{\# $\boldsymbol{Q}_i$ is the hash bucket corresponding to $\boldsymbol{q}_i$.}
\STATE \hspace{0.32cm}$\delta_i\gets \{\}$\ \textcolor{gray}{\# $\delta_i$ is the index set corresponding to $\boldsymbol{q}_i$.}
\STATE \hspace{0.32cm}\textbf{for} $j=1,...,hw$ \ \ \textcolor{gray}{\# this loop can be run in parallel.}
\STATE \hspace{0.75cm}$\boldsymbol{q}_j^{'} \gets \boldsymbol{M}\boldsymbol{q}_j$ \ \ \textcolor{gray}{\# $\boldsymbol{q}_j$ is the \emph{j}-th component of $\boldsymbol{Q}$.}
\STATE \hspace{0.75cm}\textbf{if} ${\rm{argmax}}(\boldsymbol{q}_j^{'})=={\rm index}$
\STATE \hspace{1.2cm}update $\boldsymbol{Q}_i$ by adding $\boldsymbol{q}_j$
\STATE \hspace{1.2cm}update $\delta_i$ by adding $j$
\STATE \hspace{0.32cm}\textbf{return}  $\boldsymbol{Q}_i, \delta_i$\ \textcolor{gray}{\# $\boldsymbol{Q}_i \in R^{c\times l}$ $\delta_i \in R^{l}$. $l$ is the bucket size.}

\STATE $\boldsymbol{L} \gets {\rm {Conv}}(\boldsymbol{X}),Reshape(\boldsymbol{L})$\ \ \textcolor{gray}{\# $\boldsymbol{L}\in R^{c\times hw}$.}
\STATE $\boldsymbol{V} \gets {\rm {Conv}}(\boldsymbol{X}),Reshape(\boldsymbol{V})$\ \ \textcolor{gray}{\# $\boldsymbol{V}\in R^{c\times hw}$.}
\STATE $\boldsymbol{L}_i\gets \boldsymbol{L}[\delta_i]$\ \ \textcolor{gray}{\# $\boldsymbol{L}_i\in R^{c\times l}$.}
\STATE $\boldsymbol{V}_i\gets \boldsymbol{V}[\delta_i]$\ \ \textcolor{gray}{\# $\boldsymbol{V}_i\in R^{c\times l}$.}
\STATE \textcolor{gray}{\# Dot Product Similarity Scoring (DPSS).}
\STATE $\boldsymbol{S}_f\gets \boldsymbol{Q}_i^{T}\boldsymbol{Q}_i$\ \ \textcolor{gray}{\# $\boldsymbol{S}_f\in R^{l\times l}$.}
\STATE \textcolor{gray}{\# Learnable Similarity Scoring (LSS).}
\STATE $\boldsymbol{S}_l\gets (\boldsymbol{W}_2\sigma(\boldsymbol{W}_1\boldsymbol{L}_i+\boldsymbol{b}_1)+\boldsymbol{b}_2)$\ \ \textcolor{gray}{\# $\boldsymbol{S}_l\in R^{l\times l}$.}
\STATE $\boldsymbol{S}_i\gets {\rm {Softmax}}(\boldsymbol{S}_f+\boldsymbol{S}_l)$\ \ \textcolor{gray}{\# $\boldsymbol{S}_i\in R^{l\times l}$.}
\STATE $\tilde{\boldsymbol{V}}_i\gets \boldsymbol{V}_i\boldsymbol{S}_i$\ \ \textcolor{gray}{\# $\tilde{\boldsymbol{V}}_i\in R^{c\times l}$.}
\STATE $\boldsymbol{V}[\delta_i]\gets \tilde{\boldsymbol{V}}_i$\ \ \textcolor{gray}{\# $\boldsymbol{V}\in R^{c\times hw}$.}
\STATE $\boldsymbol{V} \gets Reshape\_Inverse(\boldsymbol{V})$ \ \textcolor{gray}{\# reshape $\boldsymbol{V}$ to $R^{c\times h\times w}$.}
\STATE \textbf{Output:} $\boldsymbol{V}$
\end{algorithmic}
\end{algorithm}
\vspace{-3mm}

\section{Experiments}\label{sec:experiments}
\subsection{Datasets and Evaluation Metrics}
Following previous studies\cite{lim2017enhanced,zhang2018image,dai2019second,mei2021image}, we used 800 images from DIV2K \cite{timofte2017ntire} as training datasets. Then, we tested our model on five standard SISR benchmarks: Set5\cite{bevilacqua2012low}, Set14\cite{zeyde2010single}, B100\cite{martin2001database}, Urban100\cite{huang2015single}, and Manga109\cite{matsui2017sketch}. All the results are evaluated by SSIM\cite{wang2004image} and PSNR metrics on the Y channel in YCbCr space.
\subsection{Implementation Details}
Our final DLSN was built on a residual backbone with 10 GLAFFMs. In each GLAFFM, the number of residual blocks in LFFB was set to 4 empirically. All intermediate features have 256 channels, except those in our GLA, which have 64 channels. The last convolution layer in our DLSN has 3 filters to transfer deep features into a 3-channel RGB image. All the convolutional kernel sizes were set to $3 \times 3$.

During training, a mini-batch consists of 16 images with patch size $48\times 48$ randomly cropped from the training datasets and was augmented by random rotation of 90, 180, and 270 degrees and horizontal flipping. The mean absolute error (MAE) was used as the loss function to train our DLSN. We used ADAM optimizer\cite{kingad2015methodforstochasticoptimization} with $\beta_1= 0.9$, $\beta_2= 0.999$, and  $\epsilon=10^{-8}$ to optimize our model. In scale factor $\times 2$, the initial learning rate was set to $10^{-4}$ and reduced to half every 300 epochs until the training stops at 1500 epochs. When training our models for scale factor $\times 3$ and $\times 4$, we initialized the model parameters with pre-trained $\times 2$ network and the learning rate $10^{-4}$ was reduced to half every 50 epochs until the fine-tunning stops at 200 epochs. All our models were implemented by PyTorch and trained on Nvidia 3090 GPUs.
\subsection{Ablation Study and Analysis}
In ablation study, we trained our DLSN on DIV2K\cite{timofte2017ntire} for classical SISR with scale factor $\times 2$ and observed the best PSNR (dB) values on Set14\cite{zeyde2010single} in $5\times10^4$ iterations.
\subsubsection{Impact of LFFB and GLA}
The effects of Local Features Fusion Block (LFFB) and Global Learnable Attention (GLA) in our basic unit Global Learnable Attention-based Features Fusion Module (GLAFFM) on SR performance are shown in \cref{tab_ablation_core}.
\begin{table}[!htbp]
\centering
\caption{Ablation study on GLAFFM (including LFFB and GLA). The best and the second best results are \textbf{highlighted} and \underline{underlined}.}
\label{tab_ablation_core}
\begin{tabular}{|c|c|c|c|c|c|}
\hline
LFFB           & \XSolidBrush     & \XSolidBrush     & \Checkmark  & \Checkmark & \Checkmark  \\ \hline
GLA with $s_f(\cdot,\cdot)$    & \Checkmark  & \XSolidBrush     & \XSolidBrush     & \Checkmark & \XSolidBrush     \\ \hline
GLA with $s_f(\cdot,\cdot)+s_l(\cdot)$ & \XSolidBrush     & \Checkmark  & \XSolidBrush     & \XSolidBrush    & \Checkmark  \\ \hline
PSNR           & 33.33 & 33.49 & 33.54 & \underline{33.65} & \textbf{33.77} \\ \hline
\end{tabular}
\footnotesize{Note: $s_f(\cdot,\cdot)$ is the fixed dot product similarity scoring (see Eq.\eqref{eq_fixed_dot_product}) and $s_l(\cdot)$ is our learnable similarity scoring (see Eq.\eqref{eq_learnable_fc}).}
\end{table}\

\begin{figure}[htbp]
\centering
\begin{minipage}[]{0.35\textwidth}
\centering
	\begin{subfigure}[]{0.25\textwidth}
		\centering
		\includegraphics[width=1.1\textwidth, height=0.8\textwidth]{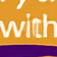}
		\vspace{-2.8mm}
	\end{subfigure}
	\hspace{1mm}
	\begin{subfigure}[]{0.25\textwidth}
		\centering
		\includegraphics[width=1.1\textwidth, height=0.8\textwidth]{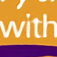}
		\vspace{-2.8mm}
	\end{subfigure}
	\hspace{1mm}
	\begin{subfigure}[]{0.25\textwidth}
		\centering
		\includegraphics[width=1.1\textwidth, height=0.8\textwidth]{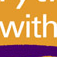}
		\vspace{-2.8mm}
	\end{subfigure}

	\begin{subfigure}[]{0.25\textwidth}
		\centering
		\includegraphics[width=1.1\textwidth, height=0.8\textwidth]{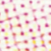}
		\vspace{-2.8mm}
	\end{subfigure}
	\hspace{1mm}
	\begin{subfigure}[]{0.25\textwidth}
		\centering
		\includegraphics[width=1.1\textwidth, height=0.8\textwidth]{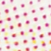}
		\vspace{-2.8mm}
	\end{subfigure}
	\hspace{1mm}
	\begin{subfigure}[]{0.25\textwidth}
		\centering
		\includegraphics[width=1.1\textwidth, height=0.8\textwidth]{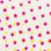}
		\vspace{-2.8mm}
	\end{subfigure}

	\begin{subfigure}[]{0.25\textwidth}
		\centering
		\includegraphics[width=1.1\textwidth, height=0.8\textwidth]{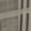}
		\subcaption*{\footnotesize Without LSS}
	\end{subfigure}
	\hspace{1mm}
	\begin{subfigure}[]{0.25\textwidth}
		\centering
		\includegraphics[width=1.1\textwidth, height=0.8\textwidth]{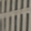}
		\subcaption*{\footnotesize With LSS}
	\end{subfigure}
	\hspace{1mm}
	\begin{subfigure}[]{0.25\textwidth}
		\centering
		\includegraphics[width=1.1\textwidth, height=0.8\textwidth]{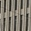}
		\subcaption*{\footnotesize HR}
	\end{subfigure}

\end{minipage}
\centering
      \caption{Comparisons between our DLSN with using our learnable similarity scoring (LSS) and without LSS for $\times 4$ SR. The textures from top to bottom belong to the 'ppt3', 'WarewareHaOniDearu' and '78004' images from Set14\cite{zeyde2010single}, Manga109\cite{matsui2017sketch} and  B100\cite{martin2001database} datasets, respectively.}
   \vspace{-3mm}
	\label{fig:fss_lss_comp}
\end{figure}

By comparing the PSNR of the first and second columns in \cref{tab_ablation_core}, we can find that our learnable similarity  scoring function $s_l(\cdot)$ can bring 0.16dB performance improvement on Set14, which is remarkable for Set14 reconstruction. Furthermore, the zoomed in results on Set14\cite{zeyde2010single}, Manga109\cite{matsui2017sketch} and  B100\cite{martin2001database} datasets for SR $\times 4$ are shown in \cref{fig:fss_lss_comp}, from which we can see that the network with using our learnable similarity scoring (LSS) can correct some inaccurate textures. Take '78004' (bottom) in B100\cite{martin2001database} as an example, our DLSN with using the LSS successfully recovers the structured architectural textures missed by the network without using the LSS. 

\begin{figure*}[htbp]
\centering
\begin{minipage}[]{\textwidth}
\rotatebox{90}{\scriptsize{Without LSS}}
\centering
	\begin{subfigure}[]{0.115\textwidth}
		\centering
		\vspace{-15mm}
		\includegraphics[width=\textwidth, height=\textwidth]{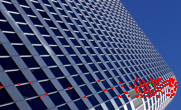}
	\end{subfigure}
	\begin{subfigure}[]{0.115\textwidth}
		\centering
		\vspace{-15mm}
		\includegraphics[width=\textwidth, height=\textwidth]{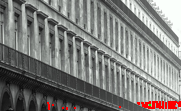}
	\end{subfigure}
	\begin{subfigure}[]{0.115\textwidth}
		\centering
		\vspace{-15mm}
		\includegraphics[width=\textwidth, height=\textwidth]{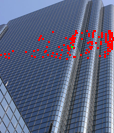}
	\end{subfigure}
	\begin{subfigure}[]{0.115\textwidth}
		\centering
		\vspace{-15mm}
		\includegraphics[width=\textwidth, height=\textwidth]{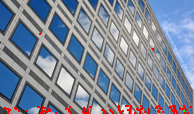}
	\end{subfigure}
	\begin{subfigure}[]{0.115\textwidth}
		\centering
		\vspace{-15mm}
		\includegraphics[width=\textwidth, height=\textwidth]{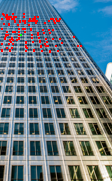}	\end{subfigure}
	\begin{subfigure}[]{0.115\textwidth}
		\centering
		\vspace{-15mm}
		\includegraphics[width=\textwidth, height=\textwidth]{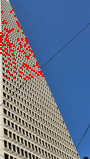}
	\end{subfigure}
	\begin{subfigure}[]{0.115\textwidth}
		\centering
		\vspace{-15mm}
		\includegraphics[width=\textwidth, height=\textwidth]{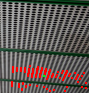}
	\end{subfigure}
\end{minipage}
\vspace{2mm}

\begin{minipage}[]{\textwidth}
\centering
	\rotatebox[origin=c]{90}{\scriptsize{With LSS}}
	\begin{subfigure}[]{0.115\textwidth}
		\centering
		\vspace{-1mm}
		\includegraphics[width=\textwidth, height=\textwidth]{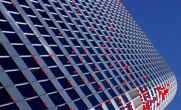}
	\end{subfigure}
	\begin{subfigure}[]{0.115\textwidth}
		\centering
		\vspace{-1mm}
		\includegraphics[width=\textwidth, height=\textwidth]{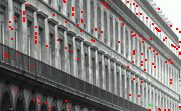}
	\end{subfigure}
	\begin{subfigure}[]{0.115\textwidth}
		\centering
		\vspace{-1mm}
		\includegraphics[width=\textwidth, height=\textwidth]{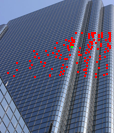}
	\end{subfigure}
	\begin{subfigure}[]{0.115\textwidth}
		\centering
		\vspace{-1mm}
		\includegraphics[width=\textwidth, height=\textwidth]{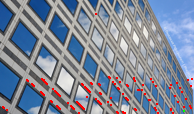}
	\end{subfigure}
	\begin{subfigure}[]{0.115\textwidth}
		\centering
		\vspace{-1mm}
		\includegraphics[width=\textwidth, height=\textwidth]{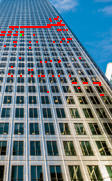}	\end{subfigure}
	\begin{subfigure}[]{0.115\textwidth}
		\centering
		\vspace{-1mm}
		\includegraphics[width=\textwidth, height=\textwidth]{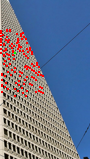}
	\end{subfigure}
	\begin{subfigure}[]{0.115\textwidth}
		\centering
		\vspace{-1mm}
		\includegraphics[width=\textwidth, height=\textwidth]{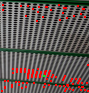}
	\end{subfigure}
\end{minipage}	
\centering
      \caption{Non-local location (red dots) comparisons between our DLSN with using learnable similarity scoring (LSS) and without LSS for x4 SR on Urban100\cite{martin2001database}. The query features are represented by green dots. Please zoom in for best view.}
	\label{fig:top_features_fss_lss_comp}
   \vspace{-4mm}
\end{figure*}

To observe the mechanism of our LSS for non-local similarity modification, we show the locations of non-local features (red dots) corresponding to the top 128 weights associated with the query feature (green dot) in the last GLA. As shown in \cref{fig:top_features_fss_lss_comp}, when the fixed dot product is used for similarity scoring (see the first line in \cref{fig:top_features_fss_lss_comp}), the associated non-local features are relatively concentrated around the query feature, that is, the fixed dot product tends to give higher weights to non-local features similar to the query feature. On the contrary, after introducing our LSS (see the second line in \cref{fig:top_features_fss_lss_comp}), an expected change occurs, our DLSN tends to assign higher weights to more informative but low-similarity non-local regions. These results indicate that our LSS can indeed correct the weights of non-local textures, and enable our DLSN to reconstruct more accurate textures by exploring non-local regions with low-similarity but richer textures information.

In \cref{tab_ablation_core}, we can see that locality inductive bias is crucial to SR performance. Without using our LFFB to explore the locality, the SR performance will degrade severely by about 0.3dB. The last column correspond to the GLAFFM used in our DLSN, which can explore local and learnable non-local information simultaneously. We also found that the fixed dot product similarity scoring function $s_f(\cdot,\cdot)$ is still helpful, and the SR performance can be further improved from 33.65 to 33.77 when composing our learnable similarity scoring function $s_l(\cdot)$ with $s_f(\cdot,\cdot)$.
\begin{table}[!htbp]
\centering
\caption{Ablation study on the number of hashing rounds.}
\label{tab_ablation_hash_number}
\begin{tabular}{|c|c|c|c|c|}
\hline
\diagbox[]{Train}{Test}    & $h$=1 & $h$=2 & $h$=3 & $h$=4 \\ \hline
$h$=1 &33.60    &33.61     &33.61     &33.61     \\ \hline
$h$=2 &33.67     &33.69    &33.73     &33.74     \\ \hline
$h$=3 &33.68     &33.76    &33.77     &33.77   \\ \hline
$h$=4 &33.68     &33.77     &33.78     &33.78    \\ \hline
\end{tabular}
   \vspace{-3mm}
\end{table}
\subsubsection{Impact of multi-hash SB-LSH and bucket size}
\noindent\textbf{Multi-hash SB-LSH.} As discussed in the \cref{sb_lsh}, increasing the number of hashing rounds can improve the robustness and effectiveness of our GLA. The number of hashing rounds $h$ can be set flexibly in the inference stage to find the trade-offs between getting accurate SR performance and reducing computational complexity. The ablation study on hashing rounds $h$ is shown in \cref{tab_ablation_hash_number}, from which we can see that increasing the number of hashing rounds in training and testing can both improve SR performance. Considering the SR performance and computational complexity, we set $h$ to be 3 in both training and testing phases in our final DLSN model.
\begin{figure}[t]
  \centering
  \includegraphics[width=0.73\linewidth]{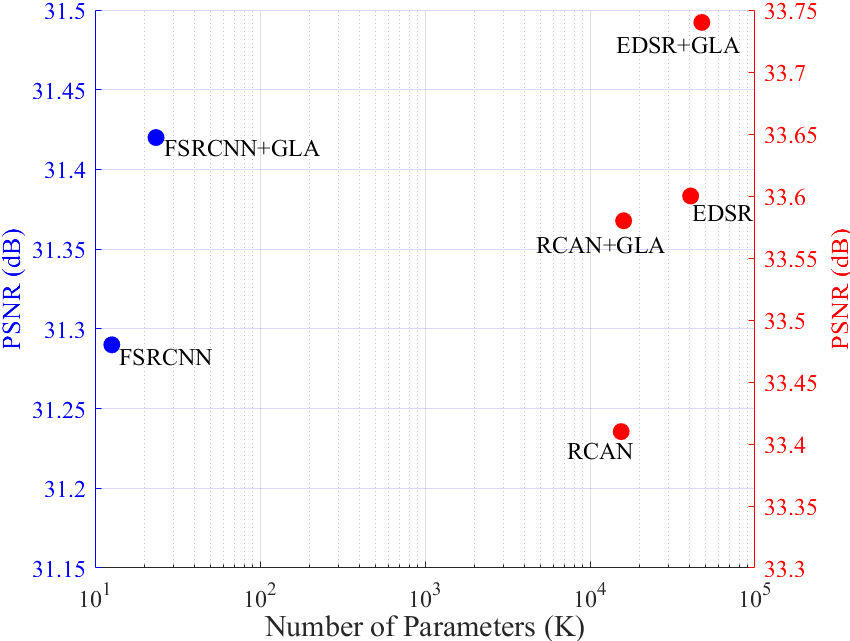}
   \caption{Parameters vs. performance. Our GLA can improve the SR performance of existing representative SISR models vary in complexity from the simple FSRCNN to the very complex EDSR and RCAN.}
   \label{fig_versatility}
\end{figure}
\begin{figure}[t]
\centering
  \includegraphics[width=0.9\linewidth,height=0.25\linewidth]{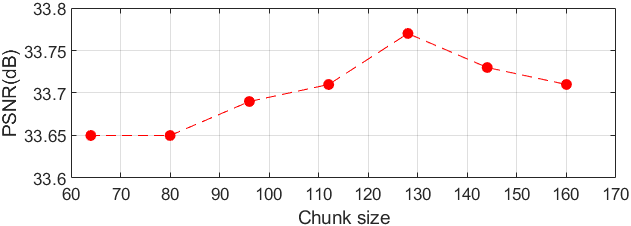}
   \caption{The PSNR results of different bucket size setting.}
   \label{bucket_size_psnr}
   \vspace{-3mm}
\end{figure}

\noindent\textbf{Bucket Size.}
As discussed in the \cref{sb_lsh}, the bucket size $l$ determine the number of non-local features that the query feature can explore. The effects of different $l$ are shown in \cref{bucket_size_psnr}, from which we can see that the SR performance of our DLSN peaks at $l=128$. When we further enlarging the bucket size, the SR performance starts to degrade. This is mainly because a larger $l$ reduces the effectiveness of our SB-LSH by merging features across multiple bucket boundaries. Adjusting the bucket size 	may further improve the reconstruction results, however, it will take a lot of training time to determine the optimal bucket size for each practical dataset. Therefore, we recommend training our DLSN with a relatively appropriate bucket size 	obtained by our ablation studies. In the extreme case when $l$ is equal to the size of the input image, our SB-LSH will be ineffective by making our GLA explore all global features.

\subsubsection{Versatility of GLA}
To analyze the versatility of our GLA, we integrate the GLA into existing representative deep SISR models with the different depths and parameters, such as FSRCNN\cite{dong2016accelerating}, EDSR\cite{lim2017enhanced}, and RCAN\cite{zhang2018image}. From \cref{fig_versatility}, we observe that our GLA can improve the SR performance of these SISR models significantly. Specifically, our GAL brings 0.13dB, 0.14dB, and 0.17dB improvement for FSRCNN\cite{dong2016accelerating}, EDSR\cite{lim2017enhanced}, and RCAN\cite{zhang2018image}, respectively. These results demonstrate that our GLA can be used as an efficient generic block to explore non-local information in deep SISR models.
\begin{figure}[t]
  \centering
  \includegraphics[width=0.6\linewidth]{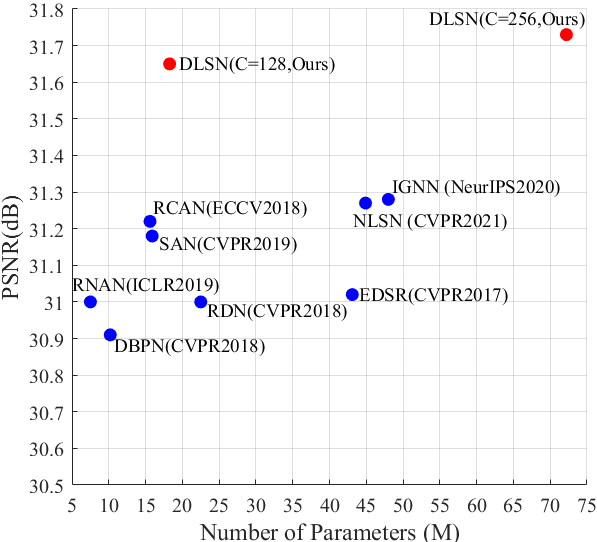}
   \caption{Model parameters and performance comparisons on Manga109 ($\times 4$).}
   \vspace{-3mm}
   \label{fig_parameters_psnr}
\end{figure}
\subsection{Efficiency Analysis}
\noindent\textbf{Model Parameters Comparisons.} We compare the model size and SR performance of our DLSN with other state-of-the-art deep SISR models including EDSR\cite{lim2017enhanced}, RDN\cite{zhang2018residual}, RCAN\cite{zhang2018image}, DBPN\cite{haris2018deep}, RNAN\cite{zhang2019residual}, SAN\cite{dai2019second}, NLSN\cite{mei2021image}, IGNN\cite{zhou2020cross}. As shown in \cref{fig_parameters_psnr}, the SR performance of our DLSN (C=256) on Manga109 ($\times 4$) is significantly better than other deep SISR models. In addition, the smaller DLSN (C=128) which can still achieve remarkable SR performance as compared to the prior state-of-the-art models. Specifically, our DLSN (C=128, about 18.27M parameters) brings 0.38dB improvement in SR performance with much lower parameters than NLSN\cite{mei2021image} (about 44.9M parameters). It means that the improvement in SR performance for our DLSN is not simply the result of having more parameters in the network.

\begin{figure}[!htbp]
\vspace{-4.7mm}
	\subfloat[]{
		   \label{fig:inference-time}
			\hspace{-2.5mm}
		\begin{minipage}[t]{0.2\textwidth}
			
			\includegraphics[height=1.1\textwidth]{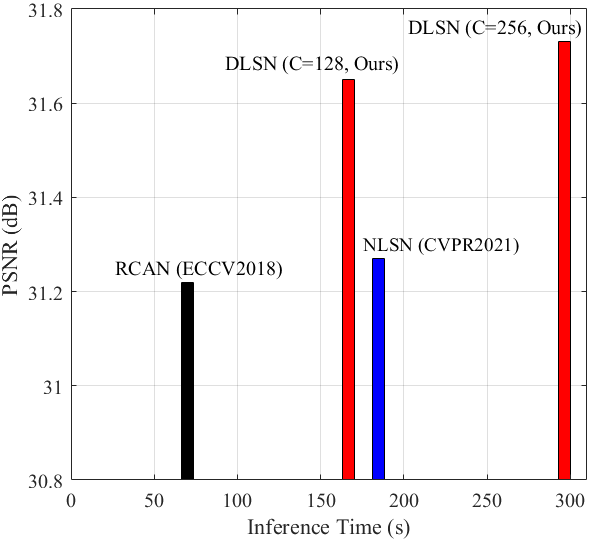}
		\end{minipage}
	}
	\hspace{4mm}
	\subfloat[]{
		   \label{fig:memory-consumption}
		\begin{minipage}[t]{0.2\textwidth}
			
			\includegraphics[height=1.1\textwidth]{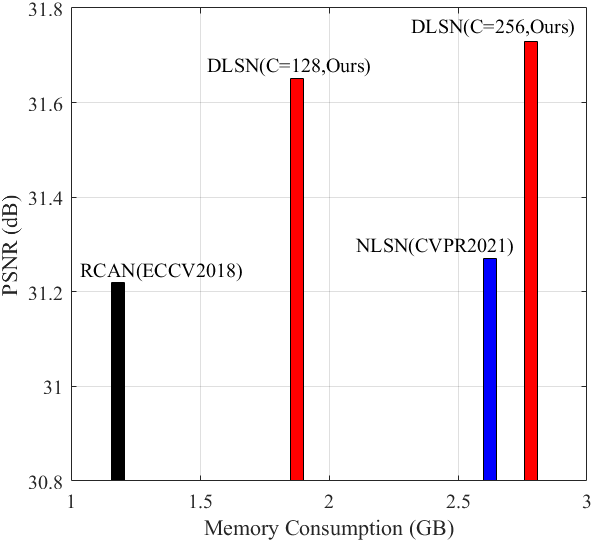}
		\end{minipage}
	}
   \caption{The PSNR results in (a) and (b) are test on Manga109 with scale factor 4. (a) Inference time comparisons on Manga109 ($\times 4$). (b) Memory consumption comparisons on an input image with size $128\times 128$.}
\vspace{-3mm}
\end{figure}

\noindent\textbf{Inference Time and Memory Comparisons.} To analysis the efficiency of our DLSN, we compare the inference time and memory consumption of recently competitive SISR models on the Manga109 dataset with scale factor 4. We also provide a smaller version of our DLSN by settting the number of channels C to 128. The inferences of all models are conducted in the same environment with Nvidia 1070Ti GPUs, and the inference time is shown in \cref{fig:inference-time}. Specifically, by comparing RCAN\cite{zhang2018image} and NLSN\cite{mei2021image}, we found that the reconstruction performance of NLSN is 0.05dB higher than that of RCAN at a cost of 114.80 seconds. Furthermore, the reconstruction performance of our DLSN (C=128) is not only 0.38dB higher than that of NLSN, but also reduces the inference time by about 18.4 seconds. These results indicate that our DLSN is very efficient in improving SR performance, and the similar conclusion can also be obtained by comparing the memory consumption (see \cref{fig:memory-consumption}). 
\begin{table*}[!htbp]
\begin{center}
\caption{Quantitative results on SISR benchmark datasets. The best and the second best results are \textbf{highlighted} and \underline{underlined}.}
\label{tab:x2_x3_x4_psnr_ssim}
\resizebox{0.8\textwidth}{8.2cm}{
\begin{tabular}{|c|c|cc|cc|cc|cc|cc|}
\hline
\multirow{2}{*}{Method} &
  \multirow{2}{*}{Scale} &
  \multicolumn{2}{c|}{Set5\cite{bevilacqua2012low}} &
  \multicolumn{2}{c|}{Set14\cite{zeyde2010single}} &
  \multicolumn{2}{c|}{B100\cite{martin2001database}} &
  \multicolumn{2}{c|}{Urban100\cite{huang2015single}} &
  \multicolumn{2}{c|}{Manga109\cite{matsui2017sketch}} \\ \cline{3-12} 
 &
   &
  \multicolumn{1}{c|}{PSNR} &
  SSIM &
  \multicolumn{1}{c|}{PSNR} &
  SSIM &
  \multicolumn{1}{c|}{PSNR} &
  SSIM &
  \multicolumn{1}{c|}{PSNR} &
  SSIM &
  \multicolumn{1}{c|}{PSNR} &
  SSIM \\ \hline

\begin{tabular}[c]{@{}c@{}}Bicubic\\ FSRCNN\cite{dong2016accelerating}\\ VDSR\cite{kim2016accurate}\\ LapSRN\cite{lai2017deep}\\ MemNet\cite{tai2017memnet}\\ SRMDNF\cite{zhang2018learning}\\ DBPN\cite{haris2018deep}\\ EDSR\cite{lim2017enhanced}\\ RDN\cite{zhang2018residual}\\ RCAN\cite{zhang2018image}\\ SAN\cite{dai2019second}\\ OISR\cite{he2019ode}\\ IGNN\cite{zhou2020cross}\\ CSNLN\cite{mei2020image}\\ HAN\cite{niu2020single}\\  NLSN\cite{mei2021image}\\ DRLN\cite{anwar2022densely}\\ SwinIR\cite{liang2021swinir}\end{tabular} &
  \begin{tabular}[c]{@{}c@{}}$\times 2$\\ $\times 2$\\ $\times 2$\\ $\times 2$\\ $\times 2$\\ $\times 2$\\ $\times 2$\\ $\times 2$\\ $\times 2$\\ $\times 2$\\ $\times 2$\\ $\times 2$\\ $\times 2$\\ $\times 2$\\ $\times 2$\\ $\times 2$\\ $\times 2$\\ $\times 2$\end{tabular} &
  \multicolumn{1}{c|}{\begin{tabular}[c]{@{}c@{}}33.66\\37.05\\37.53\\37.52\\ 37.78\\ 37.79\\ 38.09\\ 38.11\\38.24\\ 38.27\\ 38.31\\ 38.21\\ 38.24\\ 38.28\\ 38.27\\ 38.34\\ 38.27\\ 38.35\end{tabular}} &
  \begin{tabular}[c]{@{}c@{}}0.9299\\0.9560\\0.9590\\0.9591\\ 0.9597\\ 0.9601\\ 0.9600\\ 0.9602\\ 0.9614\\ 0.9614\\ 0.9620\\ 0.9612\\ 0.9613\\ 0.9616\\ 0.9614\\ 0.9618\\ 0.9616\\ 0.9620\end{tabular} &
  \multicolumn{1}{c|}{\begin{tabular}[c]{@{}c@{}}30.24\\32.66\\33.05\\33.08\\ 33.28\\ 33.32\\ 33.85\\ 33.92\\ 34.01\\ 34.12\\ 34.07\\ 33.94\\ 34.12\\ 34.07\\ 34.16\\ 34.08\\ 34.28\\ 34.14\end{tabular}} &
  \begin{tabular}[c]{@{}c@{}}0.8688\\0.9090\\0.9130\\0.9130\\ 0.9142\\ 0.9159\\ 0.9190\\ 0.9195\\ 0.9212\\ 0.9216\\ 0.9213\\ 0.9206\\ 0.9217\\ 0.9223\\ 0.9217\\ 0.9231\\ 0.9231\\ 0.9227\end{tabular} &
  \multicolumn{1}{c|}{\begin{tabular}[c]{@{}c@{}}29.56\\31.53\\31.90\\31.08\\ 32.08\\ 32.05\\ 32.27\\ 32.32\\ 32.34\\ 32.41\\ 32.42\\ 32.36\\ 32.41\\ 32.40\\ 32.41\\ 32.43\\ 32.44\\ 32.44\end{tabular}} &
  \begin{tabular}[c]{@{}c@{}}0.8431\\0.8920\\0.8960\\0.8950\\ 0.8978\\ 0.8985\\ 0.9000\\ 0.9013\\ 0.9017\\ 0.9027\\ 0.9028\\ 0.9019\\ 0.9025\\ 0.9024\\ 0.9027\\ 0.9027\\ 0.9028\\ 0.9030\end{tabular} &
  \multicolumn{1}{c|}{\begin{tabular}[c]{@{}c@{}}26.88\\29.88\\30.77\\30.41\\ 31.31\\ 31.33\\ 32.55\\ 32.93\\ 32.89\\ 33.34\\ 33.10\\ 33.03\\ 33.23\\ 33.25\\ 33.35\\ 33.42\\ 33.37\\ 33.40\end{tabular}} &
  \begin{tabular}[c]{@{}c@{}}0.8403\\0.9020\\0.9140\\0.9101\\ 0.9195\\ 0.9204\\ 0.9324\\ 0.9351\\ 0.9353\\ 0.9384\\ 0.9370\\ 0.9365\\ 0.9383\\ 0.9386\\ 0.9385\\ 0.9394\\ 0.9390\\ 0.9393\end{tabular} &
  \multicolumn{1}{c|}{\begin{tabular}[c]{@{}c@{}}30.80\\36.67\\37.22\\37.27\\ 37.72\\ 38.07\\ 38.89\\ 39.10\\ 39.18\\ 39.44\\ 39.32\\ --\\ 39.35\\ 39.37\\ 39.46\\ 39.59\\ 39.58\\  39.60\end{tabular}} &
  \begin{tabular}[c]{@{}c@{}}0.9339\\0.9710\\0.9750\\0.9740\\ 0.9740\\ 0.9761\\ 0.9775\\ 0.9773\\ 0.9780\\ 0.9786\\ 0.9792\\ --\\ 0.9786\\ 0.9785\\ 0.9785\\ 0.9789\\ 0.9786\\ 0.9792\end{tabular} \\ \hline
\begin{tabular}[c]{@{}c@{}}DLSN(ours)\\ DLSN+(ours)\end{tabular} &
  \begin{tabular}[c]{@{}c@{}}$\times 2$\\ $\times 2$\end{tabular} &
  \multicolumn{1}{c|}{\begin{tabular}[c]{@{}c@{}}\underline{38.43}\\ \textbf{38.49}\end{tabular}} &
  \begin{tabular}[c]{@{}c@{}}\underline{0.9622}\\ \textbf{0.9624}\end{tabular} &
  \multicolumn{1}{c|}{\begin{tabular}[c]{@{}c@{}}\underline{34.44}\\ \textbf{34.51}\end{tabular}} &
  \begin{tabular}[c]{@{}c@{}}\underline{0.9245}\\ \textbf{0.9251}\end{tabular} &
  \multicolumn{1}{c|}{\begin{tabular}[c]{@{}c@{}}\underline{32.46}\\ \textbf{32.53}\end{tabular}} &
  \begin{tabular}[c]{@{}c@{}}\underline{0.9036}\\ \textbf{0.9042}\end{tabular} &
  \multicolumn{1}{c|}{\begin{tabular}[c]{@{}c@{}}\underline{33.70}\\ \textbf{33.98}\end{tabular}} &
  \begin{tabular}[c]{@{}c@{}}\underline{0.9415}\\ \textbf{0.9432}\end{tabular} &
  \multicolumn{1}{c|}{\begin{tabular}[c]{@{}c@{}}\underline{39.70}\\ \textbf{39.89}\end{tabular}} &
  \begin{tabular}[c]{@{}c@{}}\underline{0.9793}\\ \textbf{0.9797}\end{tabular} \\ \hline
\begin{tabular}[c]{@{}c@{}}Bicubic\\ FSRCNN\cite{dong2016accelerating}\\ VDSR\cite{kim2016accurate}\\ LapSRN\cite{lai2017deep}\\ MemNet\cite{tai2017memnet}\\ SRMDNF\cite{zhang2018learning}\\ EDSR\cite{lim2017enhanced}\\ RDN\cite{zhang2018residual}\\ RCAN\cite{zhang2018image}\\ SAN\cite{dai2019second}\\ OISR\cite{he2019ode}\\ IGNN\cite{zhou2020cross}\\ CSNLN\cite{mei2020image}\\ HAN\cite{niu2020single}\\ NLSN\cite{mei2021image}\\ DRLN\cite{anwar2022densely}\\ SwinIR\cite{liang2021swinir}\end{tabular} &
  \begin{tabular}[c]{@{}c@{}}$\times 3$\\ $\times 3$\\ $\times 3$\\ $\times 3$\\ $\times 3$\\ $\times 3$\\ $\times 3$\\ $\times 3$\\ $\times 3$\\ $\times 3$\\ $\times 3$\\ $\times 3$\\ $\times 3$\\ $\times 3$\\ $\times 3$\\ $\times 3$\\ $\times 3$\end{tabular} &
  \multicolumn{1}{c|}{\begin{tabular}[c]{@{}c@{}}30.39 \\33.18\\33.67\\33.82\\ 34.09\\ 34.12\\ 34.65\\ 34.71\\ 34.74\\ 34.75\\ 34.72\\ 34.72\\ 34.74\\ 34.75\\ 34.85\\ 34.78\\ 34.89\end{tabular}} &
  \begin{tabular}[c]{@{}c@{}}0.8682\\0.9140\\0.9210\\0.9227\\ 0.9248\\ 0.9254\\ 0.9280\\ 0.9296\\ 0.9299\\ 0.9300\\ 0.9297\\ 0.9298\\ 0.9300\\ 0.9299\\ 0.9306\\ 0.9303\\ \underline{0.9312}\end{tabular} &
  \multicolumn{1}{c|}{\begin{tabular}[c]{@{}c@{}}27.55\\29.37\\29.78\\29.87\\ 30.00\\ 30.04\\ 30.52\\ 30.57\\ 30.65\\ 30.59\\ 30.57\\ 30.66\\ 30.66\\ 30.67\\ 30.70\\ 30.73\\ 30.77\end{tabular}} &
  \begin{tabular}[c]{@{}c@{}}0.7742\\0.8240\\0.8320\\0.8320\\ 0.8350\\ 0.8382\\ 0.8462\\ 0.8468\\ 0.8482\\ 0.8476\\ 0.8470\\ 0.8484\\ 0.8482\\  0.8483\\ 0.8485\\ 0.8488\\ 0.8503\end{tabular} &
  \multicolumn{1}{c|}{\begin{tabular}[c]{@{}c@{}}27.21\\28.53\\28.83\\28.82\\ 28.96\\ 28.97\\ 29.25\\ 29.26\\ 29.32\\ 29.33\\ 29.29\\ 29.31\\ 29.33\\ 29.32\\ 29.34\\ 29.36\\ 29.37\end{tabular}} &
  \begin{tabular}[c]{@{}c@{}}0.7385\\0.7910\\0.7990\\0.7980\\ 0.8001\\ 0.8025\\ 0.8093\\ 0.8093\\ 0.8111\\ 0.8112\\ 0.8103\\ 0.8105\\ 0.8105\\ 0.8110\\ 0.8117\\ 0.8117\\ 0.8124\end{tabular} &
  \multicolumn{1}{c|}{\begin{tabular}[c]{@{}c@{}}24.46\\26.43\\27.14\\27.07\\ 27.56\\ 27.57\\ 28.80\\ 28.80\\ 29.09\\ 28.93\\ 28.95\\ 29.03\\ 29.13\\ 29.10\\ 29.25\\ 29.21\\ 29.29\end{tabular}} &
  \begin{tabular}[c]{@{}c@{}}0.7349\\0.8080\\ 0.8290\\0.8280\\ 0.8376\\ 0.8398\\ 0.8653\\ 0.8653\\ 0.8702\\ 0.8671\\ 0.8680\\ 0.8696\\ 0.8712\\ 0.8705\\ 0.8726\\ 0.8722\\ 0.8744\end{tabular} &
  \multicolumn{1}{c|}{\begin{tabular}[c]{@{}c@{}}26.95\\31.10\\32.01\\32.21\\ 32.51\\ 33.00\\ 34.17\\ 34.13\\ 34.44\\ 34.30\\ --\\ 34.39\\ 34.45\\ 34.48\\ 34.57\\ 34.71\\ 34.74\end{tabular}} &
  \begin{tabular}[c]{@{}c@{}}0.8556\\0.9210\\0.9340\\0.9350\\ 0.9369\\ 0.9403\\ 0.9476\\ 0.9484\\ 0.9499\\ 0.9494\\ --\\ 0.9496\\ 0.9502\\ 0.9500\\ 0.9508\\  0.9509\\ 0.9518\end{tabular} \\ \hline
\begin{tabular}[c]{@{}c@{}}DLSN(ours) \\ DLSN+(ours)\end{tabular} &
  \begin{tabular}[c]{@{}c@{}}$\times 3$\\ $\times 3$\end{tabular} &
  \multicolumn{1}{c|}{\begin{tabular}[c]{@{}c@{}}\underline{34.92}\\ \textbf{35.02}\end{tabular}} &
  \begin{tabular}[c]{@{}c@{}} 0.9308\\ \textbf{0.9315}\end{tabular} &
  \multicolumn{1}{c|}{\begin{tabular}[c]{@{}c@{}}\underline{30.80}\\ \textbf{30.90}\end{tabular}} &
  \begin{tabular}[c]{@{}c@{}}\underline{0.8509}\\ \textbf{0.8521}\end{tabular} &
  \multicolumn{1}{c|}{\begin{tabular}[c]{@{}c@{}}\underline{29.41}\\ \textbf{29.47}\end{tabular}} &
  \begin{tabular}[c]{@{}c@{}} \underline{0.8136}\\ \textbf{0.8145}\end{tabular} &
  \multicolumn{1}{c|}{\begin{tabular}[c]{@{}c@{}}\underline{29.54}\\ \textbf{29.77}\end{tabular}} &
  \begin{tabular}[c]{@{}c@{}}\underline{0.8775}\\ \textbf{0.8805}\end{tabular} &
  \multicolumn{1}{c|}{\begin{tabular}[c]{@{}c@{}}\underline{34.90}\\ \textbf{35.20}\end{tabular}} &
  \begin{tabular}[c]{@{}c@{}}\underline{0.9522}\\ \textbf{0.9535}\end{tabular} \\ \hline
\begin{tabular}[c]{@{}c@{}}Bicubic\\ FSRCNN\cite{dong2016accelerating}\\ VDSR\cite{kim2016accurate}\\ LapSRN\cite{lai2017deep}\\ MemNet\cite{tai2017memnet}\\ SRMDNF\cite{zhang2018learning}\\ DBPN\cite{haris2018deep}\\ EDSR\cite{lim2017enhanced}\\ RDN\cite{zhang2018residual}\\ RCAN\cite{zhang2018image}\\ SAN\cite{dai2019second}\\ OISR\cite{he2019ode}\\ IGNN\cite{zhou2020cross}\\ CSNLN\cite{mei2020image}\\ HAN\cite{niu2020single}\\ NLSN\cite{mei2021image}\\ DRLN\cite{anwar2022densely}\\ SwinIR\cite{liang2021swinir}\end{tabular} &
  \begin{tabular}[c]{@{}c@{}}$\times 4$\\ $\times 4$\\ $\times 4$\\ $\times 4$\\ $\times 4$\\ $\times 4$\\ $\times 4$\\ $\times 4$\\ $\times 4$\\ $\times 4$\\ $\times 4$\\ $\times 4$\\ $\times 4$\\ $\times 4$\\ $\times 4$\\ $\times 4$\\ $\times 4$\\ $\times 4$\end{tabular} &
  \multicolumn{1}{c|}{\begin{tabular}[c]{@{}c@{}}28.42\\30.72\\31.35\\31.54\\ 31.74\\ 31.96\\ 32.47\\ 32.46\\ 32.47\\ 32.63\\ 32.64\\ 32.53\\ 32.57\\ 32.68\\ 32.64\\ 32.59\\ 32.63\\ 32.72\end{tabular}} &
  \begin{tabular}[c]{@{}c@{}}0.8104\\0.8660\\0.8830\\0.8850\\ 0.8893\\ 0.8925\\ 0.8980\\ 0.8968\\ 0.8990\\ 0.9002\\ 0.9003\\ 0.8992\\ 0.8998\\ 0.9004\\ 0.9002\\ 0.9000\\ 0.9002\\ \underline{0.9021}\end{tabular} &
  \multicolumn{1}{c|}{\begin{tabular}[c]{@{}c@{}}26.00\\27.61\\28.02\\28.19\\ 28.26\\ 28.35\\ 28.82\\ 28.80\\ 28.81\\ 28.87\\ 28.92\\ 28.86\\ 28.85\\ 28.95\\ 28.90\\ 28.87\\ 28.94\\ 28.94\end{tabular}} &
  \begin{tabular}[c]{@{}c@{}}0.7027\\0.7550\\0.7680\\0.7720\\ 0.7723\\ 0.7787\\ 0.7860\\ 0.7876\\ 0.7871\\ 0.7889\\ 0.7888\\ 0.7878\\ 0.7891\\ 0.7888\\ 0.7890\\ 0.7891\\ 0.7900\\ \underline{0.7914}\end{tabular} &
  \multicolumn{1}{c|}{\begin{tabular}[c]{@{}c@{}}25.96\\26.98\\27.29\\27.32\\ 27.40\\ 27.49\\ 27.72\\ 27.71\\ 27.72\\ 27.77\\ 27.78\\ 27.75\\ 27.77\\ 27.80\\ 27.80\\ 27.78\\ 27.83\\ 27.83\end{tabular}} &
  \begin{tabular}[c]{@{}c@{}}0.6675\\0.7150\\0.0726\\0.7270\\ 0.7281\\ 0.7337\\ 0.7400\\ 0.7420\\ 0.7419\\ 0.7436\\ 0.7436\\ 0.7428\\ 0.7434\\ 0.7439\\ 0.7442\\ 0.7444\\  0.7444\\ 0.7459 \end{tabular} &
  \multicolumn{1}{c|}{\begin{tabular}[c]{@{}c@{}}23.14\\24.62\\25.18\\25.21\\ 25.50\\ 25.68\\ 26.38\\ 26.64\\ 26.61\\ 26.82\\ 26.79\\ 26.79\\ 26.84\\ 27.22\\ 26.85\\ 26.96\\ 26.98\\ 27.07\end{tabular}} &
  \begin{tabular}[c]{@{}c@{}}0.6577\\0.7280\\0.7540\\0.7560\\ 0.7630\\ 0.7731\\ 0.7946\\ 0.8033\\ 0.8028\\ 0.8087\\ 0.8068\\ 0.8068\\ 0.8090\\ 0.8168\\ 0.8094\\ 0.8109\\ 0.8119\\ 0.8164\end{tabular} &
  \multicolumn{1}{c|}{\begin{tabular}[c]{@{}c@{}}24.89\\27.90\\ 28.83\\29.09\\ 29.42\\ 30.09\\ 30.91\\ 31.02\\ 31.00\\ 31.22\\ 31.18\\ --\\ 31.28\\ 31.43\\ 31.42\\ 31.27\\ 31.54\\ 31.67\end{tabular}} &
  \begin{tabular}[c]{@{}c@{}}0.7866\\0.8610\\0.8870\\0.8900\\ 0.8942\\ 0.9024\\ 0.9137\\ 0.9148\\ 0.9151\\ 0.9173\\ 0.9169\\ --\\ 0.9182\\ 0.9201\\ 0.9177\\ 0.9184\\ 0.9196\\ \underline{0.9226}\end{tabular} \\ \hline
\begin{tabular}[c]{@{}c@{}}DLSN(ours)\\ DLSN+(ours)\end{tabular} &
  \begin{tabular}[c]{@{}c@{}}$\times 4$\\ $\times 4$\end{tabular} &
  \multicolumn{1}{c|}{\begin{tabular}[c]{@{}c@{}}\underline{32.81}\\ \textbf{32.95} \end{tabular}} &
  \begin{tabular}[c]{@{}c@{}} 0.9012\\ \textbf{0.9026} \end{tabular} &
  \multicolumn{1}{c|}{\begin{tabular}[c]{@{}c@{}}\underline{29.02}\\ \textbf{29.14}\end{tabular}} &
  \begin{tabular}[c]{@{}c@{}}\underline{0.7914}\\ \textbf{0.7938}\end{tabular} &
  \multicolumn{1}{c|}{\begin{tabular}[c]{@{}c@{}}\underline{27.85}\\ \textbf{27.92}\end{tabular}} &
  \begin{tabular}[c]{@{}c@{}}\underline{0.7468}\\ \textbf{0.7483}\end{tabular} &
  \multicolumn{1}{c|}{\begin{tabular}[c]{@{}c@{}}\underline{27.26}\\ \textbf{27.49}\end{tabular}} &
  \begin{tabular}[c]{@{}c@{}}\underline{0.8191} \\ \textbf{0.8235}\end{tabular} &
  \multicolumn{1}{c|}{\begin{tabular}[c]{@{}c@{}}\underline{31.73}\\ \textbf{32.10}\end{tabular}} &
  \begin{tabular}[c]{@{}c@{}} 0.9224 \\ \textbf{0.9252}\end{tabular} \\ \hline
\end{tabular}}
\end{center}
\vspace{-3mm}
\end{table*}

\subsection{Comparisons with State-of-the-art}
\subsubsection{Bicubic-downscale degradation}
To verify the effectiveness of the proposed DLSN, we compare it with 17 state-of-the-art methods including FSRCNN\cite{dong2016accelerating}, VDSR\cite{kim2016accurate}, LapSRN\cite{lai2017deep}, EDSR\cite{lim2017enhanced}, MemNet\cite{tai2017memnet}, SRMDNF\cite{zhang2018learning}, DBPN\cite{haris2018deep}, RDN\cite{zhang2018residual}, RCAN\cite{zhang2018image}, SAN\cite{dai2019second}, OISR\cite{he2019ode},IGNN\cite{zhou2020cross}, CSNLN\cite{mei2020image}, HAN\cite{niu2020single}, NLSN\cite{mei2021image}, DRLN\cite{anwar2022densely}, and SwinIR\cite{liang2021swinir}. DLSN+ is the self ensemble results of our DLSN.
\vspace{1mm}

\noindent\textbf{Quantitative Evaluations.} The quantitative comparisons with different scale factors are shown in \cref{tab:x2_x3_x4_psnr_ssim}, where we report the quantitative comparisons between our DLSN and 17 state-of-the-art deep SISR methods on five benchmark datasets for scale factor $\times 2$, $\times 3$ and $\times 4$. From \cref{tab:x2_x3_x4_psnr_ssim}, we can see that our DLSN outperforms other state-of-the-art deep SISR models by a large margin on almost all benchmarks and scale factors. For example, compared with impressive NLSN\cite{mei2021image} in scale factor $\times 3$, our DLSN has 0.07dB, 0.10dB, 0.07dB, 0.29dB and 0.33dB performance improvement on Set5, Set14, B100, Urban100 and Manga109 datasets, respectively. The proposed DLSN can achieve significant improvements on challenging datasets Urban100 and Manga109, which contain extensive repeated textures.

On Urban100 ($\times 2$) dataset, which is designed for analyzing the self-similarity, we can see that recent deep SISR models have made very limited improvements on this dataset. For example, even though NLSN\cite{mei2021image} integrates impressive sparse non-local attention in deep SISR models for exploring the self-similarity, PSNR only increases by 0.08dB from RCAN (ECCV2018)\cite{zhang2018image} to NLSN (CVPR2021)\cite{mei2021image}. In contrast, our DLSN achieves a significant improvement on this dataset: we bring 0.36dB improvement in PSNR compared to RCAN\cite{zhang2018image}. The improvement is consistent with our motivation to design the GLA, which aims to capture the self-similarity information in LR images efficiently.
\begin{table*}[!htbp]
\centering
\caption{Quantitative results on benchmark datasets with blur-downscale degradation. The best and the second best results are \textbf{highlighted} and \underline{underlined}.}
\label{tab_x3_blur_psnr_ssim}
\begin{tabular}{|c|c|cc|cc|cc|cc|cc|}
\hline
\multirow{2}{*}{Method} &
  \multirow{2}{*}{Scale} &
  \multicolumn{2}{c|}{Set5\cite{bevilacqua2012low}} &
  \multicolumn{2}{c|}{Set14\cite{zeyde2010single}} &
  \multicolumn{2}{c|}{B100\cite{martin2001database}} &
  \multicolumn{2}{c|}{Urban100\cite{huang2015single}} &
  \multicolumn{2}{c|}{Manga109\cite{matsui2017sketch}} \\ \cline{3-12} 
 &
   &
  \multicolumn{1}{c|}{PSNR} &
  SSIM &
  \multicolumn{1}{c|}{PSNR} &
  SSIM &
  \multicolumn{1}{c|}{PSNR} &
  SSIM &
  \multicolumn{1}{c|}{PSNR} &
  SSIM &
  \multicolumn{1}{c|}{PSNR} &
  SSIM \\ \hline
\begin{tabular}[c]{@{}c@{}}Bicubic\\ FSRCNN\cite{dong2016accelerating}\\ VDSR\cite{kim2016accurate}\\ SRMDNF\cite{zhang2018learning}\\ RDN\cite{zhang2018residual}\\ EDSR\cite{lim2017enhanced}\end{tabular} &
  \begin{tabular}[c]{@{}c@{}}$\times 3$\\ $\times 3$\\ $\times 3$\\ $\times 3$\\ $\times 3$\\ $\times 3$\end{tabular} &
  \multicolumn{1}{c|}{\begin{tabular}[c]{@{}c@{}}28.78\\ 32.33\\ 33.25\\ 34.01\\ 34.58\\ 34.64\end{tabular}} &
  \begin{tabular}[c]{@{}c@{}}0.8308\\ 0.9020\\ 0.9150\\ 0.9242\\ 0.9280\\ 0.9282\end{tabular} &
  \multicolumn{1}{c|}{\begin{tabular}[c]{@{}c@{}}26.38\\ 28.91\\ 29.46\\ 30.11\\ 30.53\\ 30.54\end{tabular}} &
  \begin{tabular}[c]{@{}c@{}}0.7271\\ 0.8122\\ 0.8244\\ 0.8364\\ 0.8447\\ 0.8451\end{tabular} &
  \multicolumn{1}{c|}{\begin{tabular}[c]{@{}c@{}}26.33\\ 28.17\\ 28.57\\ 28.98\\ 29.23\\ 29.27\end{tabular}} &
  \begin{tabular}[c]{@{}c@{}}0.6918\\ 0.7791\\ 0.7893\\ 0.8009\\ 0.8079\\ 0.8094\end{tabular} &
  \multicolumn{1}{c|}{\begin{tabular}[c]{@{}c@{}}23.52\\ 25.71\\ 26.61\\ 27.50\\ 28.46\\ 28.64\end{tabular}} &
  \begin{tabular}[c]{@{}c@{}}0.6862\\ 0.7842\\ 0.8136\\ 0.8370\\ 0.8582\\ 0.8618\end{tabular} &
  \multicolumn{1}{c|}{\begin{tabular}[c]{@{}c@{}}25.46\\ 29.37\\ 31.06\\ 32.97\\ 33.97\\ 34.13\end{tabular}} &
  \begin{tabular}[c]{@{}c@{}}0.8149\\ 0.8985\\ 0.9234\\ 0.9391\\ 0.9465\\ 0.9477\end{tabular} \\
\begin{tabular}[c]{@{}c@{}}RCAN\cite{zhang2018image}\\ SAN\cite{dai2019second}\\ HAN\cite{niu2020single}\\ DLSN(Ours)\\ DLSN+(Ours)\end{tabular} &
  \begin{tabular}[c]{@{}c@{}}$\times 3$\\ $\times 3$\\ $\times 3$\\ $\times 3$\\ $\times 3$\end{tabular} &
  \multicolumn{1}{c|}{\begin{tabular}[c]{@{}c@{}}34.70\\ 34.75\\ 34.76\\ \underline{34.93}\\ \textbf{35.02}\end{tabular}} &
  \begin{tabular}[c]{@{}c@{}}0.9288\\ 0.9290\\ 0.9294\\ \underline{0.9300}\\ \textbf{0.9307}\end{tabular} &
  \multicolumn{1}{c|}{\begin{tabular}[c]{@{}c@{}}30.63\\ 30.68\\ 30.70\\ \underline{30.80}\\ \textbf{30.92}\end{tabular}} &
  \begin{tabular}[c]{@{}c@{}}0.8462\\ 0.8466\\ 0.8475\\ \underline{0.8492}\\ \textbf{0.8506}\end{tabular} &
  \multicolumn{1}{c|}{\begin{tabular}[c]{@{}c@{}}29.32\\ 29.33\\ 29.34\\ \underline{29.42}\\ \textbf{29.48}\end{tabular}} &
  \begin{tabular}[c]{@{}c@{}}0.8093\\ 0.8101\\ 0.8106 \\ \underline{0.8126}\\ \textbf{0.8136}\end{tabular} &
  \multicolumn{1}{c|}{\begin{tabular}[c]{@{}c@{}}28.81\\ 28.83\\ 28.99\\ \underline{29.38}\\ \textbf{29.62}\end{tabular}} &
  \begin{tabular}[c]{@{}c@{}}0.8647\\ 0.8646\\ 0.8676\\ \underline{0.8740}\\ \textbf{0.8772}\end{tabular} &
  \multicolumn{1}{c|}{\begin{tabular}[c]{@{}c@{}}34.38\\ 34.46\\ 34.56\\ \underline{34.98} \\ \textbf{35.29}\end{tabular}} &
  \begin{tabular}[c]{@{}c@{}}0.9483\\ 0.9487\\ 0.9494\\ \underline{0.9515}\\ \textbf{0.9529}\end{tabular} \\ \hline
\end{tabular}
\vspace{-3mm}
\end{table*}

\begin{figure}
\vspace{-1.7mm}
    \begin{minipage}[]{0.3\linewidth}
        \centering
        \includegraphics[height=1.17\linewidth,width=\linewidth]{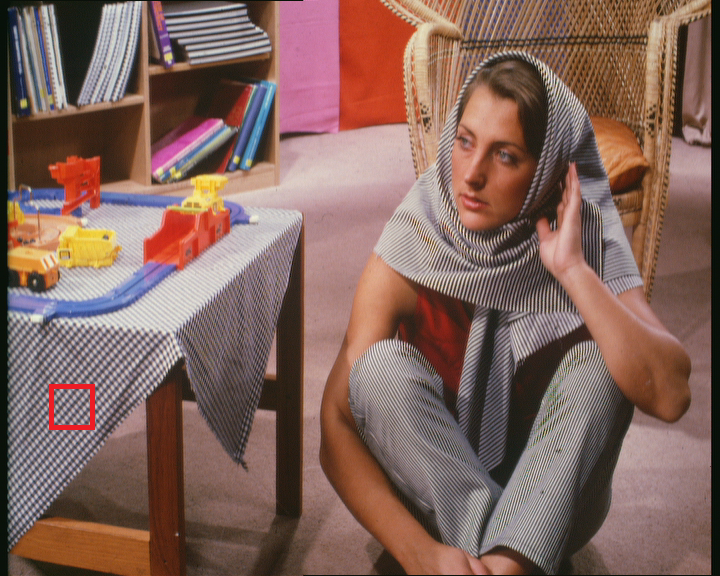}
	   \caption*{barbara}
    \end{minipage}
    \begin{minipage}[]{0.18\linewidth}
        \centering
        	
        	\begin{subfigure}[]{\linewidth}
				\centering
				\includegraphics[height=0.7\linewidth,width=\linewidth]{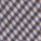}
        		\caption*{\scriptsize HR PSNR/SSIM\centering}
        		
        		\vspace{0.5mm}
        		\includegraphics[height=0.7\linewidth,width=\linewidth]{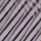}
        		\caption*{\parbox[c]{12mm}{\scriptsize EDSR\cite{lim2017enhanced} 26.72/0.8045}\centering}
			\end{subfigure}
    \end{minipage}
    \begin{minipage}[]{0.18\linewidth}
        \centering
        	
        	\begin{subfigure}[]{\linewidth}
				\centering
			\vspace{0.2mm}	\includegraphics[height=0.7\linewidth,width=\linewidth]{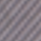}
        		\caption*{\scriptsize Bicubic 25.49/0.7104\centering}
        		
        		\vspace{0.5mm}
        		\includegraphics[height=0.7\linewidth,width=\linewidth]{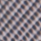}
        		\caption*{\parbox[c]{12mm}{\scriptsize HAN\cite{niu2020single} \textbf{27.58}/\underline{0.8233}}\centering}
			\end{subfigure}
    \end{minipage}
        \begin{minipage}[]{0.18\linewidth}
        \centering
        	\begin{subfigure}[]{\linewidth}
				\centering
			\vspace{1mm}
	\includegraphics[height=0.7\linewidth,width=\linewidth]{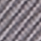}
			\vspace{-4mm}
        		\caption*{\parbox[c]{14mm}{\scriptsize FSRCNN\cite{dong2016accelerating} 26.82/0.7840}\centering}

        		\vspace{1.5mm}
        		\includegraphics[height=0.7\linewidth,width=\linewidth]{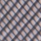}
        		\caption*{\scriptsize DLSN(ours) \underline{27.57}/\textbf{0.8244}\centering}
			\end{subfigure}
    \end{minipage}
\centering
   \caption{Visual comparisons for $\times$3 SISR with blur-downscale degradation on the Set14. The best result is \textbf{highlighted}.}
\vspace{-3mm}
   \label{fig:visual-comparison-blur-set14}
\end{figure}

\begin{figure*}[t]
\centering
\hspace{13mm}\begin{minipage}[]{0.20\textwidth}
\vspace{-2mm}
\begin{subfigure}[]{1.0\textwidth}
\centering
\includegraphics[height=1.02\textwidth,width=1\textwidth]{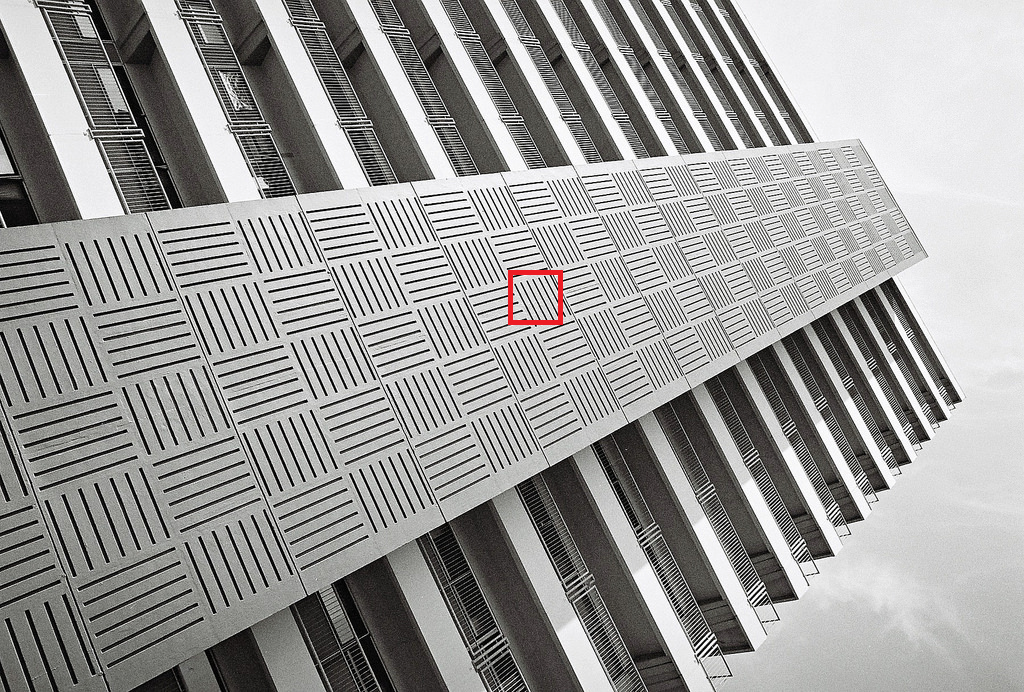}
\caption*{img\_092 \centering}
\end{subfigure}

\vspace{3mm}
\begin{subfigure}[]{1.0\textwidth}
\centering
\includegraphics[height=1.02\textwidth,width=1\textwidth]{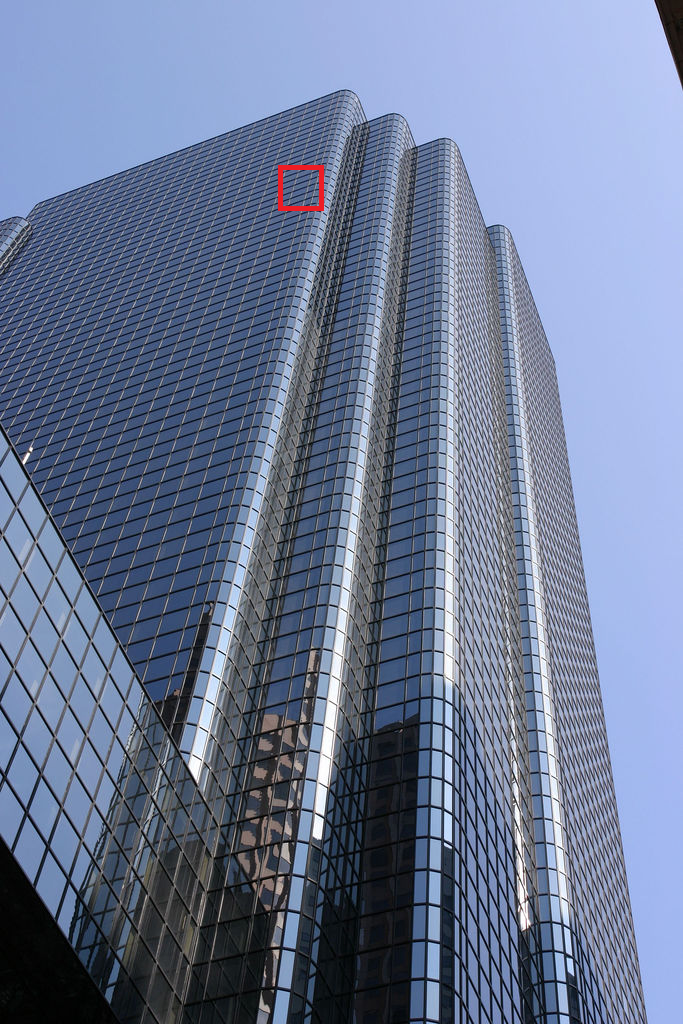}
\caption*{img\_074 \centering}
\end{subfigure}
\end{minipage}
\hspace{-12mm}
\captionsetup[subfigure]{justification=justified,singlelinecheck=false}
\begin{minipage}[]{0.75\textwidth}
\centering
\begin{subfigure}[]{0.15\textwidth}
		\centering
		\includegraphics[width=1.1\textwidth,height=0.7\textwidth]{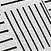}
		\caption*{\parbox[c]{6mm}{\footnotesize HR PSNR/SSIM}}
	\end{subfigure}
\hspace{1mm}
	\begin{subfigure}[]{0.15\textwidth}
		\centering
		\includegraphics[width=1.1\textwidth,height=0.7\textwidth]{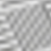}
		\caption*{\parbox[c]{8mm}{\footnotesize Bicubic 16.58/0.4375}}
	\end{subfigure}
\hspace{1mm}
	\begin{subfigure}[]{0.15\textwidth}
		\centering
		\includegraphics[width=1.1\textwidth,height=0.7\textwidth]{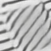}
		\caption*{\parbox[c]{16mm}{\footnotesize LapSRN\cite{lai2017deep} 18.20/0.6077}}
	\end{subfigure}
\hspace{1mm}
	\begin{subfigure}[]{0.15\textwidth}
		\centering
		\includegraphics[width=1.1\textwidth,height=0.7\textwidth]{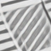}
		\caption*{\parbox[c]{13mm}{\footnotesize EDSR\cite{lim2017enhanced}  19.15/0.6785}}
	\end{subfigure}
	\hspace{1mm}
	\begin{subfigure}[]{0.15\textwidth}
		\centering
		\includegraphics[width=1.1\textwidth,height=0.7\textwidth]{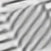}
		\caption*{\parbox[c]{13mm}{\footnotesize VDSR\cite{kim2016accurate} 18.14/0.6012}}
	\end{subfigure}

\vspace{0.8mm}
	\begin{subfigure}[]{0.15\textwidth}
		\centering
		\includegraphics[width=1.1\textwidth,height=0.7\textwidth]{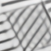}
		\caption*{\parbox[c]{12mm}{\footnotesize RDN\cite{zhang2018residual} 19.17/0.6769}}
	\end{subfigure}
\hspace{1mm}
	\begin{subfigure}[]{0.15\textwidth}
		\centering
		\includegraphics[width=1.1\textwidth,height=0.7\textwidth]{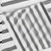}
		\caption*{\parbox[c]{14mm}{\footnotesize RCAN\cite{zhang2018image} 19.64/0.6962}}
	\end{subfigure}
\hspace{1mm}
	\begin{subfigure}[]{0.15\textwidth}
		\centering
		\includegraphics[width=1.1\textwidth,height=0.7\textwidth]{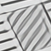}
		\caption*{\parbox[c]{11mm}{\footnotesize SAN\cite{dai2019second} 19.60/0.6987}}
	\end{subfigure}
\hspace{1mm}
	\begin{subfigure}[]{0.15\textwidth}
		\centering
		\includegraphics[width=1.1\textwidth,height=0.7\textwidth]{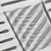}
		\caption*{\parbox[c]{14mm}{\footnotesize NLSN\cite{mei2021image} 19.51/0.6922}}
	\end{subfigure}
\hspace{1mm}
	\begin{subfigure}[]{0.15\textwidth}
		\centering
		\vspace{-0.2mm}
		\includegraphics[width=1.1\textwidth,height=0.7\textwidth]{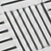}
		\caption*{\parbox[c]{10mm}{\footnotesize \textbf{DLSN(Ours) 20.25/0.7235}}}
	\end{subfigure}

\vspace{1mm}
\begin{subfigure}[]{0.15\textwidth}
		\centering
		\includegraphics[width=1.1\textwidth,height=0.7\textwidth]{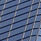}
		\caption*{\parbox[c]{6mm}{\footnotesize HR PSNR/SSIM}}
	\end{subfigure}
\hspace{1mm}
	\begin{subfigure}[]{0.15\textwidth}
		\centering
		\includegraphics[width=1.1\textwidth,height=0.7\textwidth]{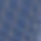}
		\caption*{\parbox[c]{8mm}{\footnotesize Bicubic 22.15/0.5552}}
	\end{subfigure}
\hspace{1mm}
	\begin{subfigure}[]{0.15\textwidth}
		\centering
		\includegraphics[width=1.1\textwidth,height=0.7\textwidth]{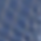}
		\caption*{\parbox[c]{16mm}{\footnotesize LapSRN\cite{lai2017deep} 23.14/0.6520}}
	\end{subfigure}
\hspace{1mm}
	\begin{subfigure}[]{0.15\textwidth}
		\centering
		\includegraphics[width=1.1\textwidth,height=0.7\textwidth]{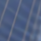}
		\caption*{\parbox[c]{13mm}{\footnotesize EDSR\cite{lim2017enhanced}  24.21/0.7345}}
	\end{subfigure}
\hspace{1mm}
	\begin{subfigure}[]{0.15\textwidth}
		\centering
		\includegraphics[width=1.1\textwidth,height=0.7\textwidth]{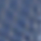}
		\caption*{\parbox[c]{13mm}{\footnotesize VDSR\cite{kim2016accurate} 23.08/0.6414}}
	\end{subfigure}

\vspace{1mm}
	\begin{subfigure}[]{0.15\textwidth}
		\centering
		\includegraphics[width=1.1\textwidth,height=0.7\textwidth]{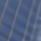}
		\caption*{\parbox[c]{12mm}{\footnotesize RDN\cite{zhang2018residual} 24.29/0.7445}}
	\end{subfigure}
\hspace{1mm}
	\begin{subfigure}[]{0.15\textwidth}
		\centering
		\includegraphics[width=1.1\textwidth,height=0.7\textwidth]{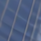}
		\caption*{\parbox[c]{14mm}{\footnotesize RCAN\cite{zhang2018image} 24.44/0.7570}}
	\end{subfigure}
\hspace{1mm}
	\begin{subfigure}[]{0.15\textwidth}
		\centering
		\includegraphics[width=1.1\textwidth,height=0.7\textwidth]{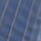}
		\caption*{\parbox[c]{10mm}{\footnotesize SAN\cite{dai2019second} 24.38/0.7479}}
	\end{subfigure}
\hspace{1mm}
	\begin{subfigure}[]{0.15\textwidth}
		\centering
		\includegraphics[width=1.1\textwidth,height=0.7\textwidth]{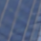}
		\caption*{\parbox[c]{14mm}{\footnotesize NLSN\cite{mei2021image} 24.98/0.7776}}
	\end{subfigure}
\hspace{1mm}
	\begin{subfigure}[]{0.15\textwidth}
		\centering
		\vspace{-0.2mm}
		\includegraphics[width=1.1\textwidth,height=0.7\textwidth]{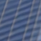}
		\caption*{\parbox[c]{10mm}{\footnotesize \textbf{DLSN(Ours) 25.23/0.7965}}}
	\end{subfigure}
\end{minipage}
\centering
   \caption{Visual comparisons on Urban100\cite{huang2015single} with scale factor 4.}
\vspace{-3mm}
   \label{fig:visual-comparison-u100}
\end{figure*}

\begin{figure*}[!htbp]
\centering
\hspace{12mm}\begin{minipage}[]{0.2\textwidth}
\begin{subfigure}[]{1.0\textwidth}
\centering
\vspace{-1.4mm}
\includegraphics[height=1.02\textwidth,width=1\textwidth]{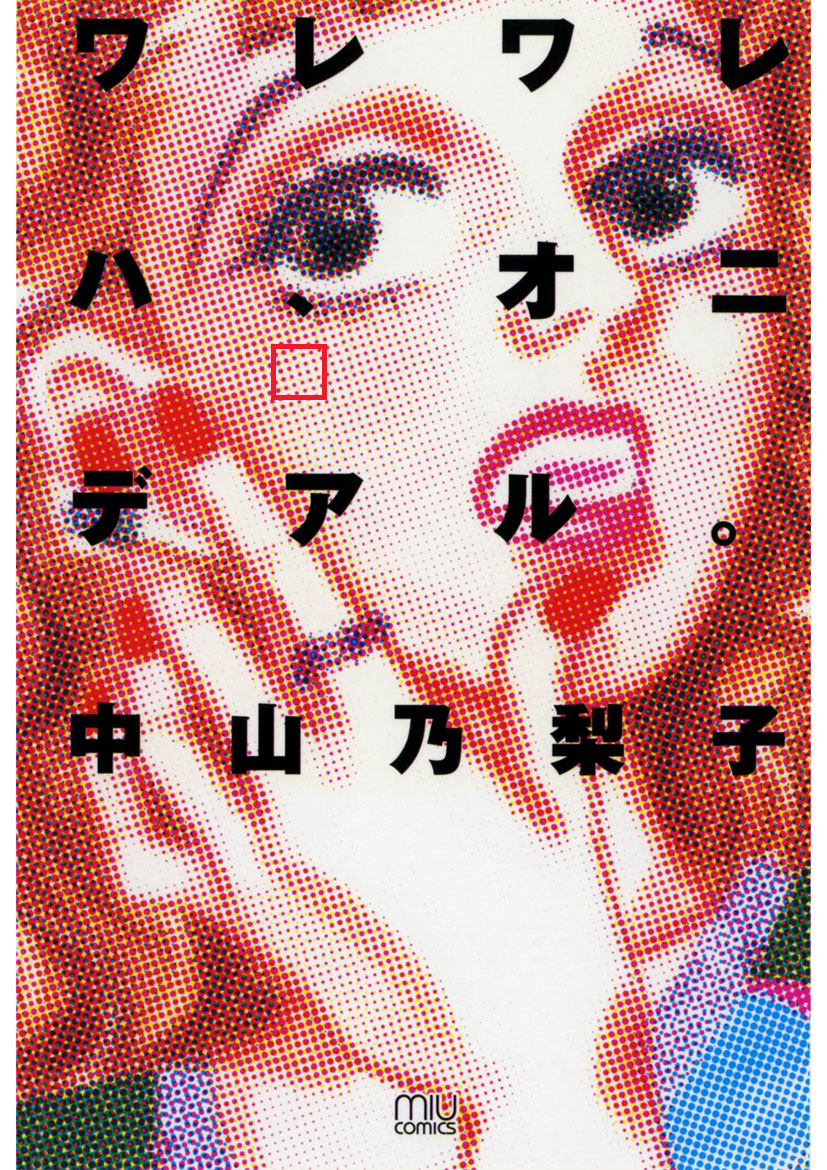}
\caption*{\footnotesize WarewareHaOniDearu \centering}
\end{subfigure}

\vspace{3.6mm}
\begin{subfigure}[]{1.0\textwidth}
\centering
\includegraphics[height=1.03\textwidth,width=1\textwidth]{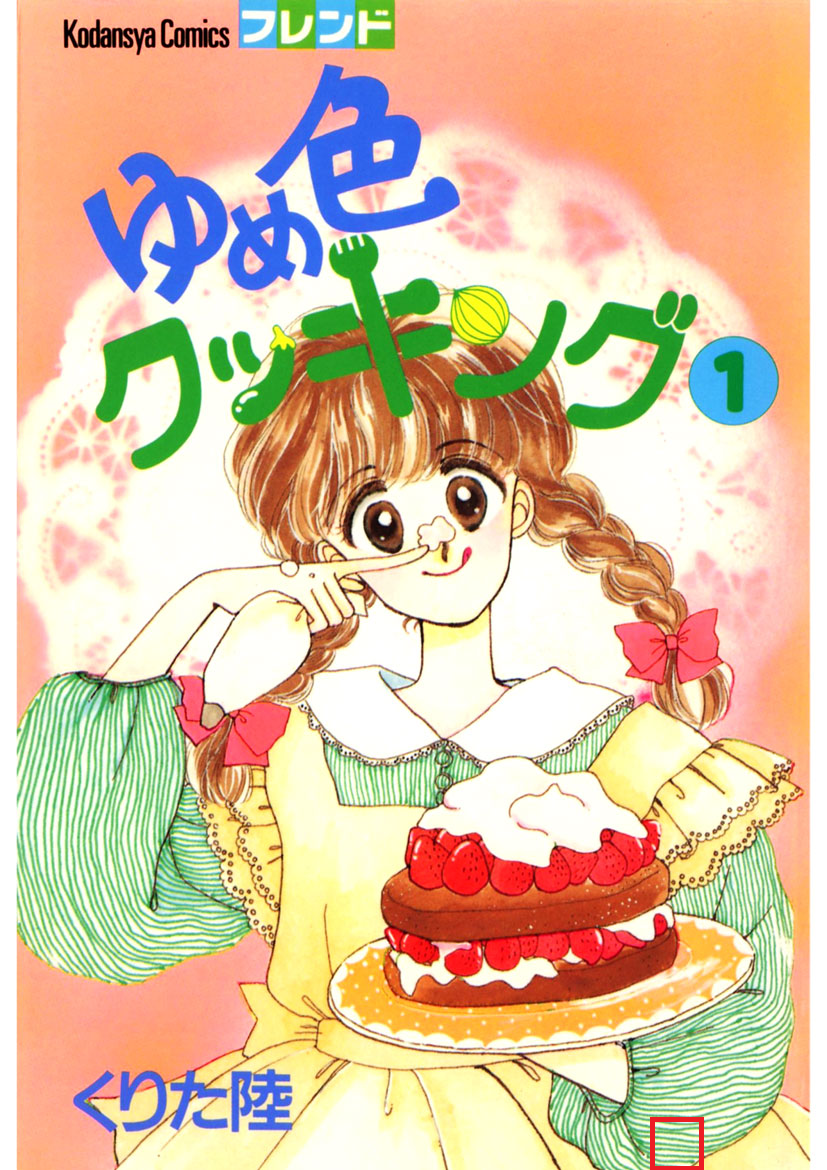}
\caption*{\footnotesize YumeiroCooking \centering}
\end{subfigure}
\end{minipage}
\hspace{-12.5mm}
\captionsetup[subfigure]{justification=justified,singlelinecheck=false}
\begin{minipage}[]{0.75\textwidth}
\centering
	\begin{subfigure}[]{0.15\textwidth}
		\centering
		\includegraphics[width=1.1\textwidth,height=0.7\textwidth]{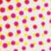}
		\caption*{\parbox[c]{8mm}{\footnotesize HR PSNR/SSIM}}
	\end{subfigure}
\hspace{1mm}
	\begin{subfigure}[]{0.15\textwidth}
		\centering
		\includegraphics[width=1.1\textwidth,height=0.7\textwidth]{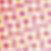}
		\caption*{\parbox[c]{8mm}{\footnotesize Bicubic 20.31/0.6101}}
	\end{subfigure}
\hspace{1mm}
	\begin{subfigure}[]{0.15\textwidth}
		\centering
		\includegraphics[width=1.1\textwidth,height=0.7\textwidth]{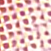}
		\caption*{\parbox[c]{16mm}{\footnotesize LapSRN\cite{lai2017deep} 22.05/0.7887}}
	\end{subfigure}
\hspace{1mm}
	\begin{subfigure}[]{0.15\textwidth}
		\centering
		\includegraphics[width=1.1\textwidth,height=0.7\textwidth]{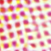}
		\caption*{\parbox[c]{13mm}{\footnotesize EDSR\cite{lim2017enhanced} 23.69/0.8424}}
	\end{subfigure}
\hspace{1mm}
	\begin{subfigure}[]{0.15\textwidth}
		\centering
		\includegraphics[width=1.1\textwidth,height=0.7\textwidth]{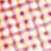}
		\caption*{\parbox[c]{16mm}{\footnotesize FSRCNN\cite{dong2016accelerating} 22.52/0.7735}}
	\end{subfigure}

\vspace{1mm}
	\begin{subfigure}[]{0.15\textwidth}
		\centering
		\includegraphics[width=1.1\textwidth,height=0.7\textwidth]{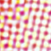}
		\caption*{\parbox[c]{12mm}{\footnotesize RDN\cite{zhang2018residual} 20.61/0.7401}}
	\end{subfigure}
\hspace{1mm}
	\begin{subfigure}[]{0.15\textwidth}
		\centering
		\includegraphics[width=1.1\textwidth,height=0.7\textwidth]{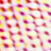}
		\caption*{\parbox[c]{14mm}{\footnotesize RCAN\cite{zhang2018image} 20.84/0.7505}}
	\end{subfigure}
\hspace{1mm}
	\begin{subfigure}[]{0.15\textwidth}
		\centering
		\includegraphics[width=1.1\textwidth,height=0.7\textwidth]{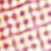}
		\caption*{\parbox[c]{14mm}{\footnotesize DRCN\cite{kim2016deeply} 22.37/0.7703}}
	\end{subfigure}
\hspace{1mm}
	\begin{subfigure}[]{0.15\textwidth}
		\centering
		\includegraphics[width=1.1\textwidth,height=0.7\textwidth]{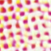}
		\caption*{\parbox[c]{14mm}{\footnotesize NLSN\cite{mei2021image} 23.00/0.8344}}
	\end{subfigure}
\hspace{1mm}
	\begin{subfigure}[]{0.15\textwidth}
		\centering
		\vspace{-0.2mm}
		\includegraphics[width=1.1\textwidth,height=0.7\textwidth]{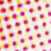}
		\caption*{\parbox[c]{15mm}{\footnotesize \textbf{DLSN(Ours) 24.83/0.8811}}}
	\end{subfigure}

\vspace{1mm}
\begin{subfigure}[]{0.15\textwidth}
		\centering
		\includegraphics[width=1.1\textwidth,height=0.7\textwidth]{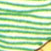}
		\caption*{\parbox[c]{8mm}{\footnotesize HR PSNR/SSIM}}
	\end{subfigure}
\hspace{1mm}
	\begin{subfigure}[]{0.15\textwidth}
		\centering
		\includegraphics[width=1.1\textwidth,height=0.7\textwidth]{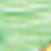}
		\caption*{\parbox[c]{8mm}{\footnotesize Bicubic 24.66/0.7861}}
	\end{subfigure}
\hspace{1mm}
	\begin{subfigure}[]{0.15\textwidth}
		\centering
		\includegraphics[width=1.1\textwidth,height=0.7\textwidth]{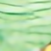}
		\caption*{\parbox[c]{16mm}{\footnotesize LapSRN\cite{lai2017deep} 26.92/0.8752}}
	\end{subfigure}
\hspace{1mm}
	\begin{subfigure}[]{0.15\textwidth}
		\centering
		\includegraphics[width=1.1\textwidth,height=0.7\textwidth]{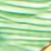}
		\caption*{\parbox[c]{13mm}{\footnotesize EDSR\cite{lim2017enhanced} 29.05/0.9243}}
	\end{subfigure}
\hspace{1mm}
	\begin{subfigure}[]{0.15\textwidth}
		\centering
		\includegraphics[width=1.1\textwidth,height=0.7\textwidth]{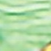}
		\caption*{\parbox[c]{16mm}{\footnotesize FSRCNN\cite{dong2016accelerating} 26.34/0.8474}}
	\end{subfigure}

\vspace{1mm}
	\begin{subfigure}[]{0.15\textwidth}
		\centering
		\includegraphics[width=1.1\textwidth,height=0.7\textwidth]{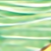}
		\caption*{\parbox[c]{12mm}{\footnotesize RDN\cite{zhang2018residual} 28.24/0.9121}}
	\end{subfigure}
\hspace{1mm}
	\begin{subfigure}[]{0.15\textwidth}
		\centering
		\includegraphics[width=1.1\textwidth,height=0.7\textwidth]{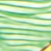}
		\caption*{\parbox[c]{14mm}{\footnotesize RCAN\cite{zhang2018image} 29.86/0.9369}}
	\end{subfigure}
\hspace{1mm}
	\begin{subfigure}[]{0.15\textwidth}
		\centering
		\includegraphics[width=1.1\textwidth,height=0.7\textwidth]{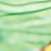}
		\caption*{\parbox[c]{14mm}{\footnotesize DRCN\cite{kim2016deeply} 26.90/0.8727}}
	\end{subfigure}
\hspace{1mm}
	\begin{subfigure}[]{0.15\textwidth}
		\centering
		\includegraphics[width=1.1\textwidth,height=0.7\textwidth]{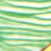}
		\caption*{\parbox[c]{14mm}{\footnotesize NLSN\cite{mei2021image} 29.57/0.9328}}
	\end{subfigure}
\hspace{1mm}
	\begin{subfigure}[]{0.15\textwidth}
		\centering
		\vspace{-0.3mm}
		\includegraphics[width=1.1\textwidth,height=0.7\textwidth]{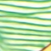}
		\caption*{\parbox[c]{15mm}{\footnotesize \textbf{DLSN(Ours) 30.01/0.9377}}}
	\end{subfigure}
\end{minipage}
\centering
   \caption{Visual comparisons on Manga109\cite{matsui2017sketch} with scale factor 4.}
\vspace{-3mm}
   \label{fig:visual-comparison-m109}
\end{figure*}


\noindent\textbf{Qualitative Evaluations.} Visual comparisons on Urban100 and Mange109 with scale factor $\times 4$ are shown in \cref{fig:visual-comparison-u100} and \cref{fig:visual-comparison-m109}, from both of which we can see that the proposed DLSN can restore the severely damaged textures when the corresponding non-local information can be found in LR images. On the contrary, deep SISR models without non-local attention cannot reconstruct severely damaged textures accurately. For example, by comparing the reconstruction results of image 'img\_092' in \cref{fig:visual-comparison-u100}, we observe that the generated results of our DLSN is very close to the HR, but the other very competitive deep SISR models without non-local attention such as EDSR\cite{lim2017enhanced}, RCAN\cite{zhang2018image} are not suitable for recovering such severely damaged regions. Moreover, compared with the other non-local deep SISR methods such as SAN\cite{dai2019second} and NLSN\cite{mei2021image}, our DLSN still maintains better reconstruction quality with more accurate textures. These comparison shows that our DLSN is more efficient in repairing severely damaged regions by exploring the self-similarity information with the proposed non-local method GLA.

The visual results demonstrate that our DLSN not only outperforms in quantitative metrics, but also perceptually better than the other deep SISR models by a large margin. In addition, more visual comparisons between our DLSN and some significant SISR methods including DRLN\cite{anwar2022densely} and SwinIR\cite{liang2021swinir} can be found in our supplementary file.
\subsubsection{Blur-downscale degradation}
Our DLSN is verified on blur-downscale degradation SISR tasks with scale factor $\times 3$ and the gaussian standard deviation is set to 1.6 as discussed in SRMDNF\cite{zhang2018learning} and RCAN\cite{zhang2018image}. The quantitative results of our DLSN are compared with the other 8 state-of-the-art methods: FSRCNN\cite{dong2016accelerating}, VDSR\cite{kim2016accurate}, EDSR\cite{lim2017enhanced}, SRMDNF\cite{zhang2018learning}, RDN\cite{zhang2018residual}, RCAN\cite{zhang2018image}, SAN\cite{dai2019second} and HAN\cite{niu2020single}.

\noindent\textbf{Quantitative Evaluations.} 
The quantitative comparisons of blur-downscale degradation with scale factor $\times 3$ are shown in \cref{tab_x3_blur_psnr_ssim}. From \cref{tab_x3_blur_psnr_ssim}, we can see that our DLSN outperforms the very deep SISR model HAN\cite{niu2020single}, which has a well-engineered structure with the holistic attention. Specifically, compared with HAN, our DLSN has 0.17dB, 0.10dB, 0.08dB, 0.39dB and 0.42dB improvement on Set5, Set14, B100, Urban100 and Manga109 datasets, respectively. These results mean that our DLSN can still achieve very impressive SR performance when solving SISR tasks with blur-downscale degradation.
\begin{table}[t]
\centering
\caption{Quantitative results on Set14 \cite{zeyde2010single} ($\times 3$) with noise-downscale degradation in $5\times10^4$ iterations. The best and the second best results are \textbf{highlighted} and \underline{underlined}.}
\label{tab_x3_noise_psnr_ssim}
\resizebox{0.49\textwidth}{!}{
\begin{tabular}{|ccccc|}
\hline
\multirow{2}{*}{Noise level} & FSRCNN\cite{dong2016accelerating}       & EDSR\cite{lim2017enhanced}        & RCAN\cite{zhang2018image}          & DLSN(Ours)            \\ \cline{2-5} 
                             & PSNR/SSIM    & PSNR/SSIM    & PSNR/SSIM    & PSNR/SSIM             \\ \hline
10                           & 27.26/0.7396 & \underline{28.77}/\underline{0.7892} & 28.76/0.7883 & \textbf{28.91}/\textbf{0.7903} \\
15                           & 26.58/0.7123 & \underline{28.11}/\underline{0.7653} & 28.06/0.7622 & \textbf{28.20}/\textbf{0.7671} \\
20                           & 25.99/0.6875 & \underline{27.52}/\underline{0.7442} & 27.48/0.7415 & \textbf{27.62}/\textbf{0.7463} \\
25                           & 25.51/0.6674 & \underline{26.99}/\underline{0.7241} & \underline{26.99}/0.7232 & \textbf{27.09}/\textbf{0.7279} \\ \hline
\end{tabular}
}
\vspace{-3mm}
\end{table}

\noindent\textbf{Qualitative Evaluations.} 
Visual comparisons on Set14 datasets with blur-downscale degradation are shown in \cref{fig:visual-comparison-blur-set14}, from which we can see that our DLSN generates the most visual pleasing textures with accurate image details. From \cref{fig:visual-comparison-blur-set14}, we can also find that EDSR\cite{lim2017enhanced} cannot restore textures which are severely damaged by blur-downscale degradation, even though the selected region has informative repeated textures in the tablecloth of the input LR image. These visual results show that our DLSN is indeed effective when dealing with blur-downscale SISR tasks.

\subsubsection{Noisy-downscale degradation}
To verify the robustness of our DLSN in handling noisy-downscale degradation, we reimplemented some state-of-the-art deep SISR models under the same training datasets and compared them with our DLSN. \cref{tab_x3_noise_psnr_ssim} shows the SR performance of these state-of-the-art SISR models at noise levels of 10, 15, 20, and 25, respectively. From \cref{tab_x3_noise_psnr_ssim}, we can see that our DLSN outperforms other state-of-the-art SISR models at all noise levels, which indicates that our DLSN is robust in handling SISR tasks with different noise levels.
\subsubsection{Reference-based SISR}
We also provided the comparisons with some significant reference-based SISR methods, including MASA\cite{lu2021masa}. Compared with reference-based SISR methods, our DLSN can still achieve competitive reconstruction performance without using the reference images. Please refer to the supplementary file for detailed comparisons.

\subsubsection{Real-world images super-resolution}
\begin{figure}[t]
	\hspace{5.5mm}
    \begin{minipage}[]{0.4\linewidth}
        \centering
	   \vspace{1.5mm}
        \includegraphics[height=0.78\linewidth,width=\linewidth]{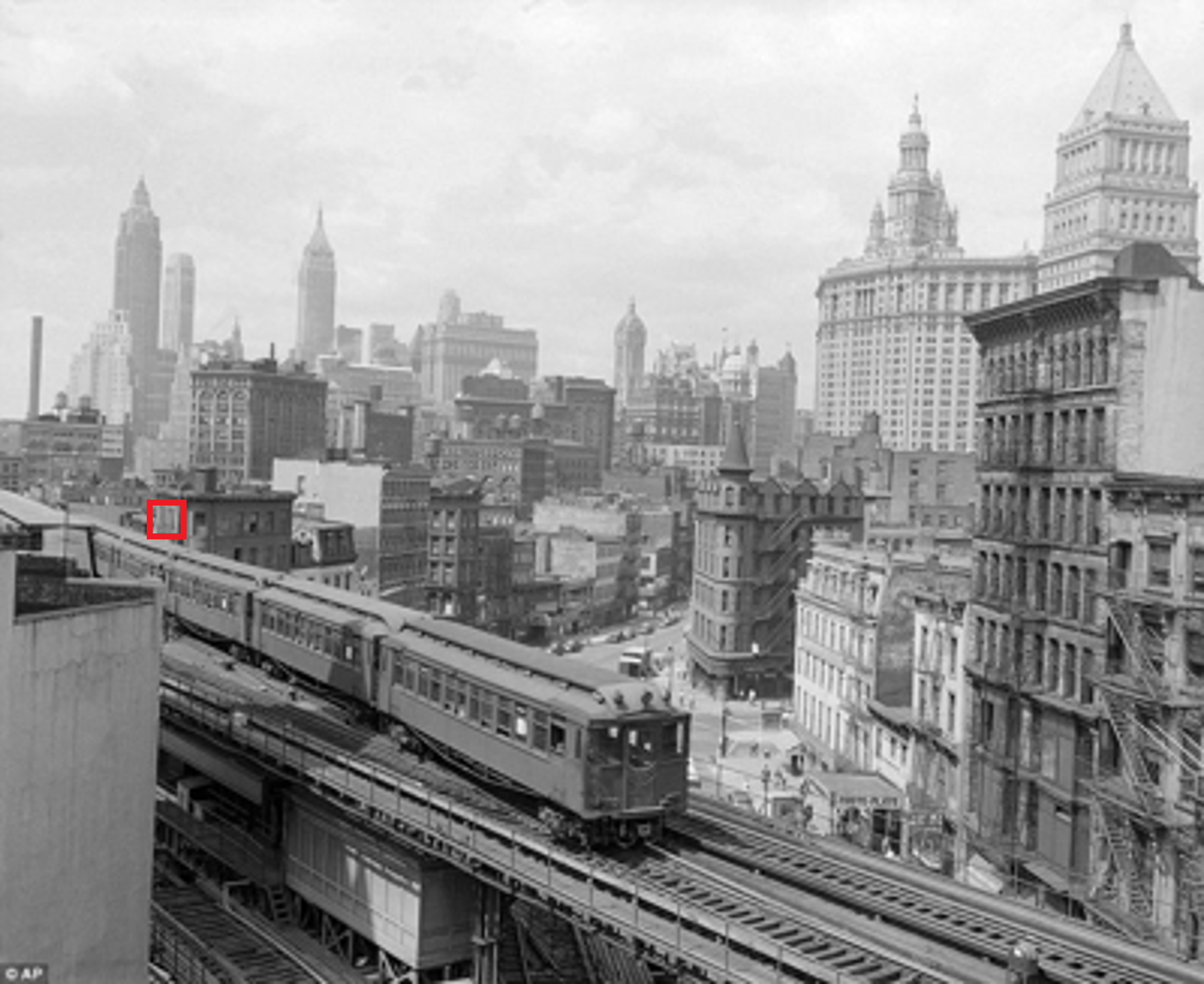}
        \caption*{\footnotesize LR input}
    \end{minipage}
    \begin{minipage}[]{0.2\linewidth}
        \centering
        	\begin{subfigure}[]{\linewidth}
				\centering
				\includegraphics[height=0.6\linewidth,width=1.13\linewidth]{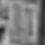}
        		\caption*{Bicubic}
        		
        		\vspace{0.5mm}
        		\includegraphics[height=0.6\linewidth,width=1.13\linewidth]{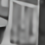}
        		\caption*{\footnotesize DRLN\cite{anwar2022densely}}
			\end{subfigure}
    \end{minipage}
    \hspace{1mm}
    \begin{minipage}[]{0.2\linewidth}
        \centering
        	
        	\begin{subfigure}[]{\linewidth}
				\centering
	\includegraphics[height=0.6\linewidth,width=1.13\linewidth]{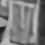}
        		\caption*{\parbox[c]{21mm}{\footnotesize MS-LapSRN\cite{lai2018fast}}}
        		
        		\vspace{0.5mm}
        	\includegraphics[height=0.6\linewidth,width=1.13\linewidth]{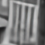}
        		\caption*{\footnotesize DLSN(Ours)}
			\end{subfigure}
    \end{minipage}

\vspace{0.5mm}
\hspace{-2.2mm}
    \begin{minipage}[]{0.4\linewidth}
        \centering
	   \vspace{1.3mm}
        \includegraphics[height=0.77\linewidth,width=1\linewidth]{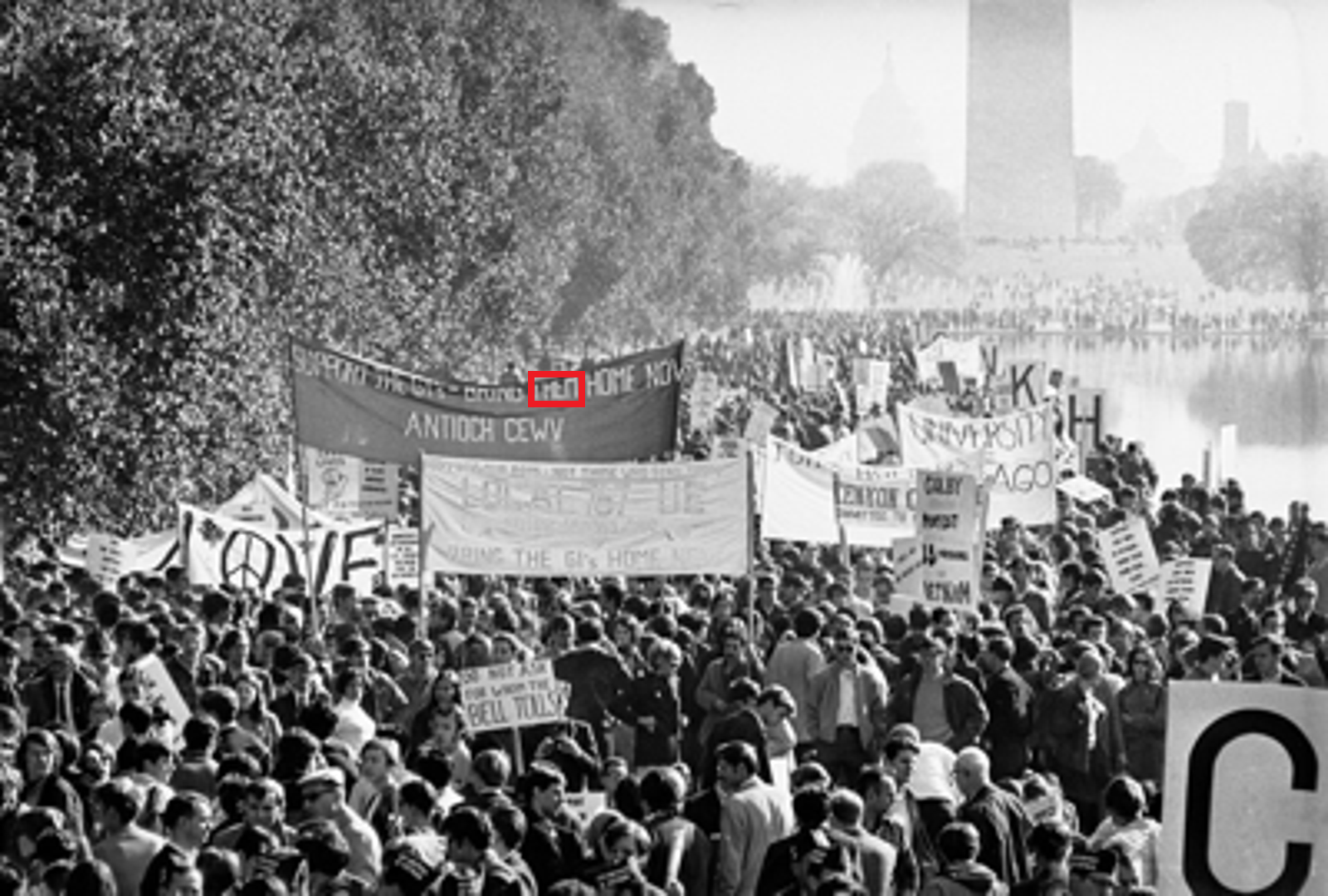}
        \caption*{\footnotesize LR input}
    \end{minipage}
    \begin{minipage}[]{0.2\linewidth}
        \centering
        	
        	\begin{subfigure}[]{\linewidth}
				\centering
				\includegraphics[height=0.6\linewidth,width=1.13\linewidth]{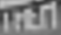}
        		\caption*{Bicubic}
        		\vspace{0.5mm}
        		\includegraphics[height=0.6\linewidth,width=1.13\linewidth]{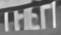}
        		\caption*{\footnotesize DRLN\cite{anwar2022densely}}
			\end{subfigure}
    \end{minipage}
    \hspace{1mm}
    \begin{minipage}[]{0.2\linewidth}
        \centering
        	
        	\begin{subfigure}[]{\linewidth}
				\centering
				\includegraphics[height=0.6\linewidth,width=1.13\linewidth]{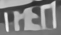}
        		\caption*{\parbox[c]{21mm}{\footnotesize MS-LapSRN\cite{lai2018fast}}}
        		\vspace{0.5mm}
        		\includegraphics[height=0.6\linewidth,width=1.13\linewidth]{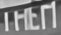}
        		\caption*{\footnotesize DLSN(Ours)}
			\end{subfigure}
    \end{minipage}
    \centering
   \caption{Visual comparisons on real-world images for scale factor 4. On the bottom LR input, Our DLSN recovers the letter“H”appropriately while DRLN and MS-LapSRN incorrectly connect the stroke with the letter “E”.}
\vspace{-3mm}
   \label{fig_real_world_cmp}
\end{figure}

In this section, we provided the performance comparisons on real-world historic images with JPEG compression artifacts, as discussed in MS-LapSRN\cite{lai2018fast} and DRLN\cite{anwar2022densely}. From \cref{fig_real_world_cmp}, we can see that on the top LR input, our DLSN reconstructs more accurate structured architectural textures than the competitive MS-LapSRN and DRLN. Furthermore, compared with the results of Bicubic, we can observe that when the input letters are seriously damaged, our DLSN can still generate sharp and clear edges of the letters.

\subsection{Limitations}
\begin{figure}[t]
    \begin{minipage}[]{0.3\linewidth}
        \centering
	\vspace{1.4mm}
        \includegraphics[height=1.07\linewidth,width=\linewidth]{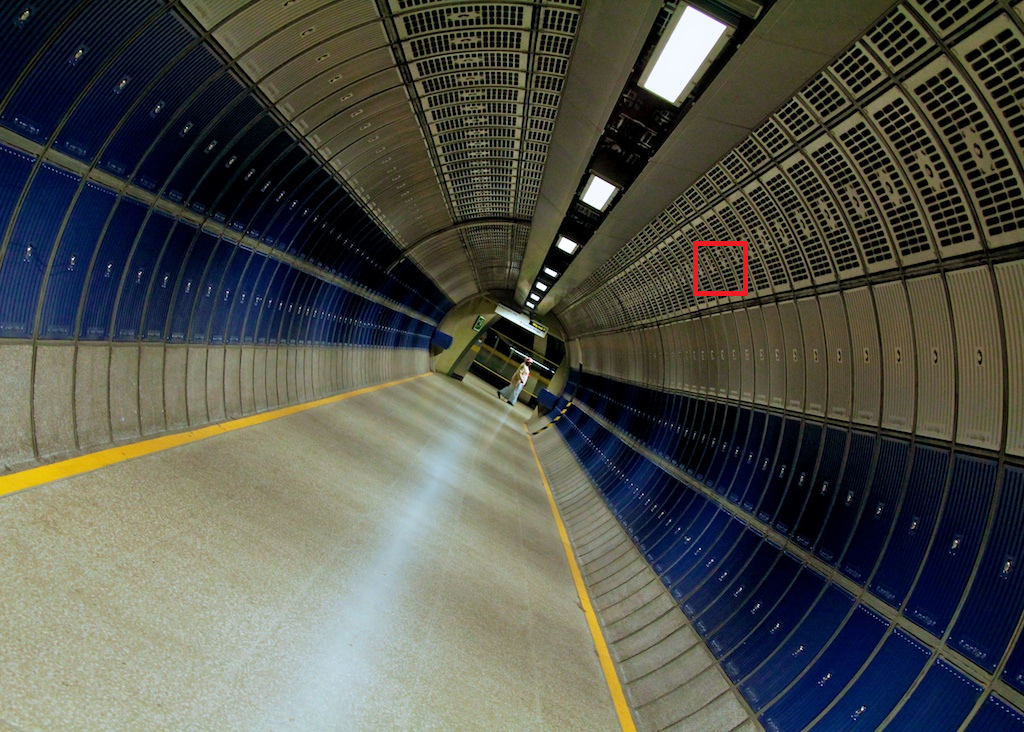}
        \caption*{\footnotesize LR input}
    \end{minipage}
    \begin{minipage}[]{0.18\linewidth}
        \centering
        	
        	\begin{subfigure}[]{\linewidth}
				\centering
				\includegraphics[height=0.7\linewidth,width=\linewidth]{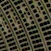}
        		\caption*{\footnotesize HR}
        		
        		\vspace{0.5mm}
        		\includegraphics[height=0.7\linewidth,width=\linewidth]{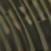}
        		\caption*{\footnotesize NLSN\cite{mei2021image}}
			\end{subfigure}
    \end{minipage}
    \begin{minipage}[]{0.18\linewidth}
        \centering
        	
        	\begin{subfigure}[]{\linewidth}
				\centering
				\includegraphics[height=0.7\linewidth,width=\linewidth]{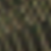}
        		\caption*{Bicubic}
        		
        		\vspace{0.5mm}
        		\includegraphics[height=0.7\linewidth,width=\linewidth]{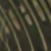}
        		\caption*{\footnotesize DRLN\cite{anwar2022densely}}
			\end{subfigure}
    \end{minipage}
        \begin{minipage}[]{0.18\linewidth}
        \centering
        	
        	\begin{subfigure}[]{\linewidth}
				\centering
	\includegraphics[height=0.7\linewidth,width=\linewidth]{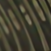}
        		\caption*{\footnotesize RCAN\cite{zhang2018image}}
        		
        		\vspace{0.5mm}
        		\includegraphics[height=0.7\linewidth,width=\linewidth]{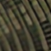}
        		\caption*{\footnotesize DLSN(ours)}
			\end{subfigure}
    \end{minipage}
    \centering
   \caption{Limitation. A failure case for repairing completely destroyed regions. Our method is not able to hallucinate details if the input regions lack the basic texture patterns for the reconstruction.}
\vspace{-3mm}
   \label{fig_limitation}
\end{figure}

Although our DLSN can reconstruct visually pleasing results, it struggles to "hallucinate" fine details if the input textures are completely destroyed. This is a common limitation shared by classic deep SISR methods including RCAN\cite{zhang2018image}, NLSN\cite{mei2021image} and DRLN\cite{anwar2022densely}. For example, as shown in \cref{fig_limitation}, all the classic deep SISR methods failed to repair the completely destroyed structured architectural region. The main reason for the limitation is that the completely destroyed input region lacks the basic texture patterns for the reconstruction.

\section{Conclusion}
In this paper, we provided new insights into the self-similarity in SISR tasks and found some critical limitations presenting in the existing deep self-similarity-based methods. To overcome these drawbacks, we design a flexible global learnable attention-based features fusion module (GLAFFM) that can make our deep learnable similarity network (DLSN) focus on more valuable non-local textures to repair severely damaged regions. Furthermore, with the super-bit locality-sensitive hashing (SB-LSH), our GLAFFM can achieve asymptotic linear computational complexity with respect to the image size when computing non-local attention. In addition, extensive experiments demonstrate that our GLA can handle SISR tasks with different degradation types (e.g. blur and noise), and can be integrated as an efficient general building block in deep SISR models.

\bibliographystyle{IEEEtranS}
\bibliography{egbib}

\vspace{-13mm}
\begin{IEEEbiography}[{\includegraphics[width=1in,height=1.25in,clip]{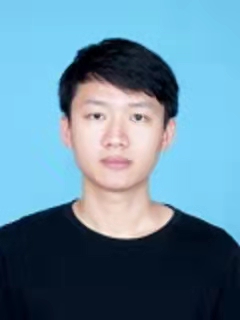}}]%
{Jian-Nan Su} received the B.S. and M.S. degree in computer science and engineering from Fuzhou University, Fuzhou, China, in 2015 and 2018. He is currently pursuing the Ph.D. degree with the College of Computer and Data Science, Fuzhou University, Fuzhou, China. His current research interests include image processing and machine learning.
\end{IEEEbiography}
\vspace{-14mm}
\begin{IEEEbiography}[{\includegraphics[width=1in,height=1.35in,clip]{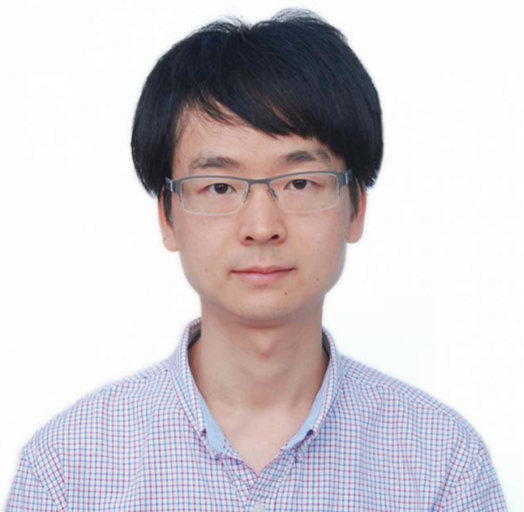}}]%
{Min Gan} received the B.S. degree in computer science and engineering from the Hubei University of Technology, Wuhan, China, in 2004, and the Ph.D. degree in control science and engineering from Central South University, Changsha, China, in 2010. He is currently a Professor with the College of Computer and Data Science, Fuzhou University, Fuzhou, China. His current research interests include statistical learning, system identification, and nonlinear time-series analysis, image processing.
\end{IEEEbiography}
\vspace{-12mm}
\begin{IEEEbiography}[{\includegraphics[width=1in,height=1.25in,clip]{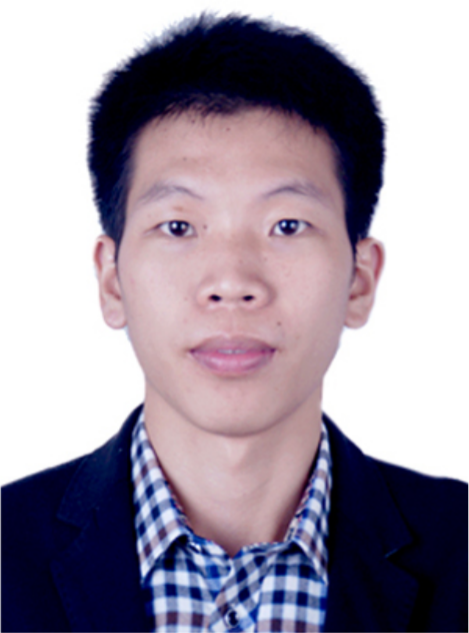}}]%
{Guang-Yong Chen} received the B.S. degree in mathematics from Xidian University, Xi’an, China, in 2012, and the M.S. degree in mathematics from the University of Science and Technology of China, Hefei, China, in 2014, and the Ph.D. degree in mathematics from Fuzhou University, Fuzhou, China, in 2019. His current research interests include computational intelligence, image processing, system identification, and nonlinear time-series analysis.
\end{IEEEbiography}
\vspace{-15mm}
\begin{IEEEbiography}[{\includegraphics[width=1in,height=1.25in,clip]{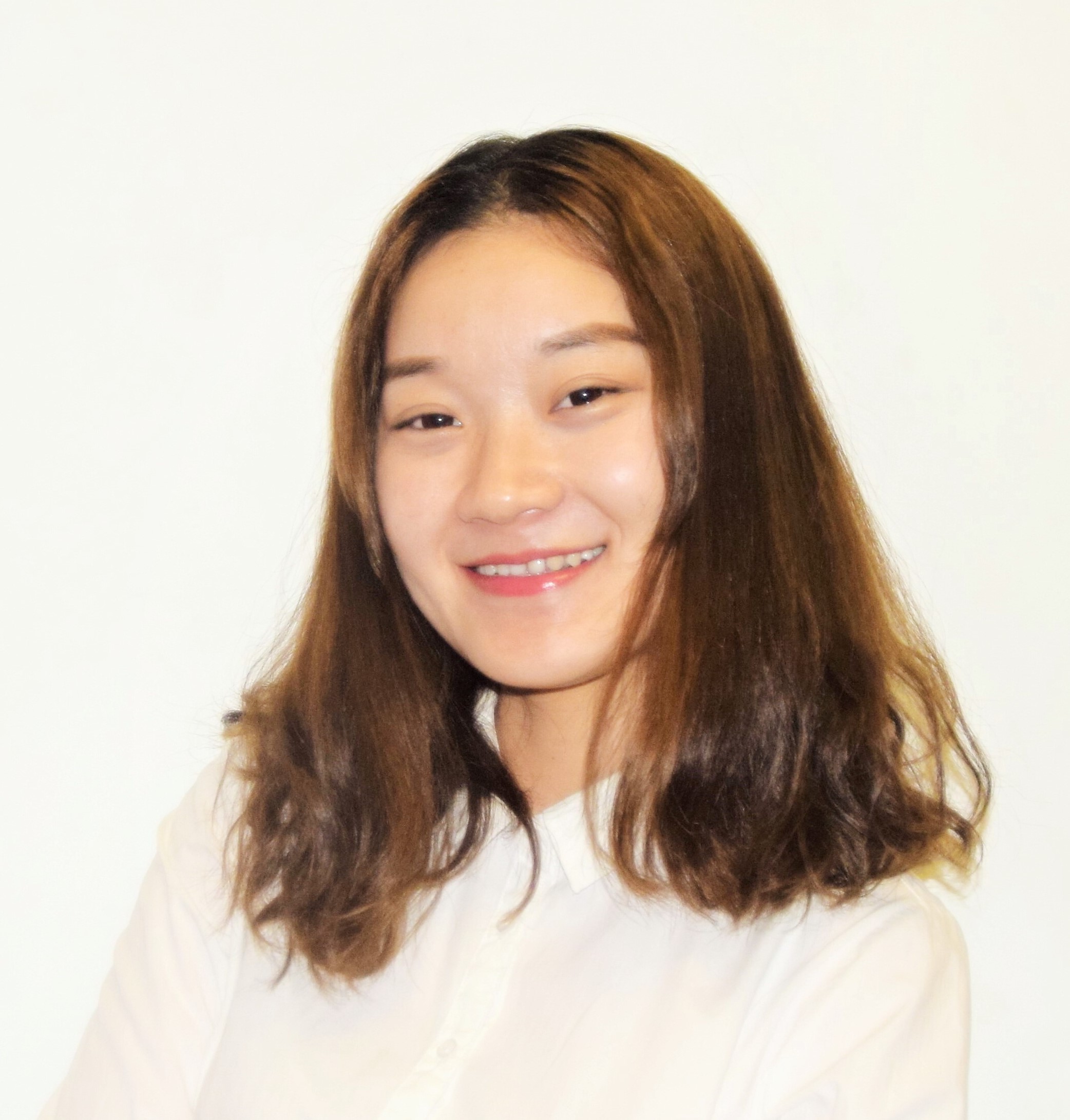}}]%
{Jia-Li Yin} received the Ph.D. degree in the Department of Computer Science and Engineering, Yuan Ze University, Taoyuan, Taiwan, in 2020. She is currently a Professor and Qishan Scholar with the College of Computer Science and Big Data, Fuzhou University, China. Her research interests include digital image processing, computer vision, pattern recognition, and deep learning.
\end{IEEEbiography}
\vspace{-12.5mm}
\begin{IEEEbiography}[{\includegraphics[width=1in,height=1.25in,clip]{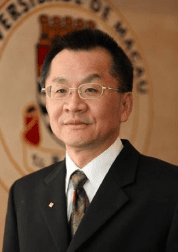}}]%
{C. L. Philip Chen} received the M.S. degree in electrical engineering from the University of Michigan, Ann Arbor, MI, USA, in 1985, and the Ph.D. degree in electrical engineering from Purdue University, West Lafayette, IN, USA, in 1988. He is currently the Dean of the School of Computer Science and Engineering, South China University of Technology, Guangzhou 510641, China. His current research interests include systems, cybernetics, and computational intelligence. Dr. Chen is a Fellow of the AAAS.
\end{IEEEbiography}
\end{document}